\documentclass[letterpaper]{article} % DO NOT CHANGE THIS
\usepackage[preprint]{aaai2027}  % Preprint version with author information
\usepackage[hyphens]{url}  % DO NOT CHANGE THIS
\usepackage{graphicx} % DO NOT CHANGE THIS
\usepackage{natbib}  % DO NOT CHANGE THIS AND DO NOT ADD ANY OPTIONS TO IT
\usepackage{caption} % DO NOT CHANGE THIS AND DO NOT ADD ANY OPTIONS TO IT
\usepackage{algorithm}
\usepackage{algorithmic}
\usepackage{pifont}
\newcommand{\cmark}{\ding{51}}
\usepackage{newfloat}
\usepackage{listings}
\DeclareCaptionStyle{ruled}{labelfont=normalfont,labelsep=colon,strut=off} % DO NOT CHANGE THIS
\floatstyle{ruled}
\newfloat{listing}{tb}{lst}{}
\floatname{listing}{Listing}

\usepackage{booktabs}
\usepackage{amsmath}
\usepackage{multirow}

\title{HarmTrace: Anchor-Calibrated Decoupled Optimization for Fine-Grained Target Identification in Harmful Memes}
\author{
    Yujia Li\textsuperscript{\rm 1},
    Yiqun Zhang\textsuperscript{\rm 2},
    Zihan Cheng\textsuperscript{\rm 1},
    Yijie Huang\textsuperscript{\rm 1},
    Tenglong Ye\textsuperscript{\rm 1},\\
    Zihan Wang\textsuperscript{\rm 1},
    Xiaocui Yang\textsuperscript{\rm 1},
    Shi Feng\textsuperscript{\rm 1}\corresponding,
    Yifei Zhang\textsuperscript{\rm 1},
    Daling Wang\textsuperscript{\rm 1}
}
\affiliations{
    \textsuperscript{\rm 1}School of Computer Science and Engineering, Northeastern University\\
    Shenyang 110819, China\\
    \textsuperscript{\rm 2}Apple Inc., Beijing 100006, China\\
    liyujia@mails.neu.edu.cn, fengshi@cse.neu.edu.cn
}

\begin{document}

\maketitle

\begin{abstract}
Multimodal harmful meme detection is typically formulated as image--text
harmfulness classification. A model may correctly predict harmfulness while misidentifying the attacked
target or its supporting evidence. We therefore extend harmful meme detection with fine-grained target
identification, asking what type of target is attacked, who is targeted, and
where the target appears in the meme. The model predicts harmfulness for every
meme and, for harmful memes, outputs the target category, target entity,
textual mention, and visual region. To support this task, we introduce \textbf{Meme3W}, which unifies multiple public harmful meme datasets and provides human-verified annotations for harmful instances.
We further
introduce \textbf{Joint Record Accuracy} (JRA), a strict record-level metric requiring the
harmfulness label and all target-identification fields to be jointly correct.
Experiments with representative multimodal large language
models reveal a substantial gap between harmfulness accuracy and JRA. 
To narrow this gap, we propose \textbf{HarmTrace}, an anchor-calibrated
decoupled optimization framework. HarmTrace strengthens target-entity supervision through entity-aware
supervised fine-tuning. It then applies Conditional Target-identification
Policy Optimization (CTPO) to decouple harmfulness and target-identification
advantages, restricting target-identification optimization to label-correct
responses for harmful examples. CTPO uses a Virtual Positive Anchor (VPA) as a
fully correct reference for target-identification advantage normalization.
HarmTrace improves both JRA and harmfulness accuracy across the evaluated backbones, 
with JRA on the Qwen3-VL-8B backbone increasing from 17.58\% to 52.51\%.
Our code is publicly available at
\url{https://github.com/llly1234/HarmTrace-for-Harmful-Memes}.

\end{abstract}

\section{Introduction}

Most existing methods formulate multimodal harmful meme detection primarily
as image--text harmfulness classification
\cite{Kmainasi2026CanTM,Cheng2026DRHMDT,Wang2026SGoTR1SG,
hou-etal-2026-beyond}.
However, a correct harmfulness label alone does not establish whether the
attacked target has been correctly identified
\cite{mia-fahim-2025-banhateme}.
It also does not indicate which textual mention and visual region support that
target identification. We therefore extend harmful meme detection with
\textbf{fine-grained target identification}, predicting harmfulness for every
meme and, for harmful memes, additionally identifying the target category,
target entity, textual mention, and visual region.

\begin{figure}[t]
\centering
\includegraphics[width=\columnwidth]{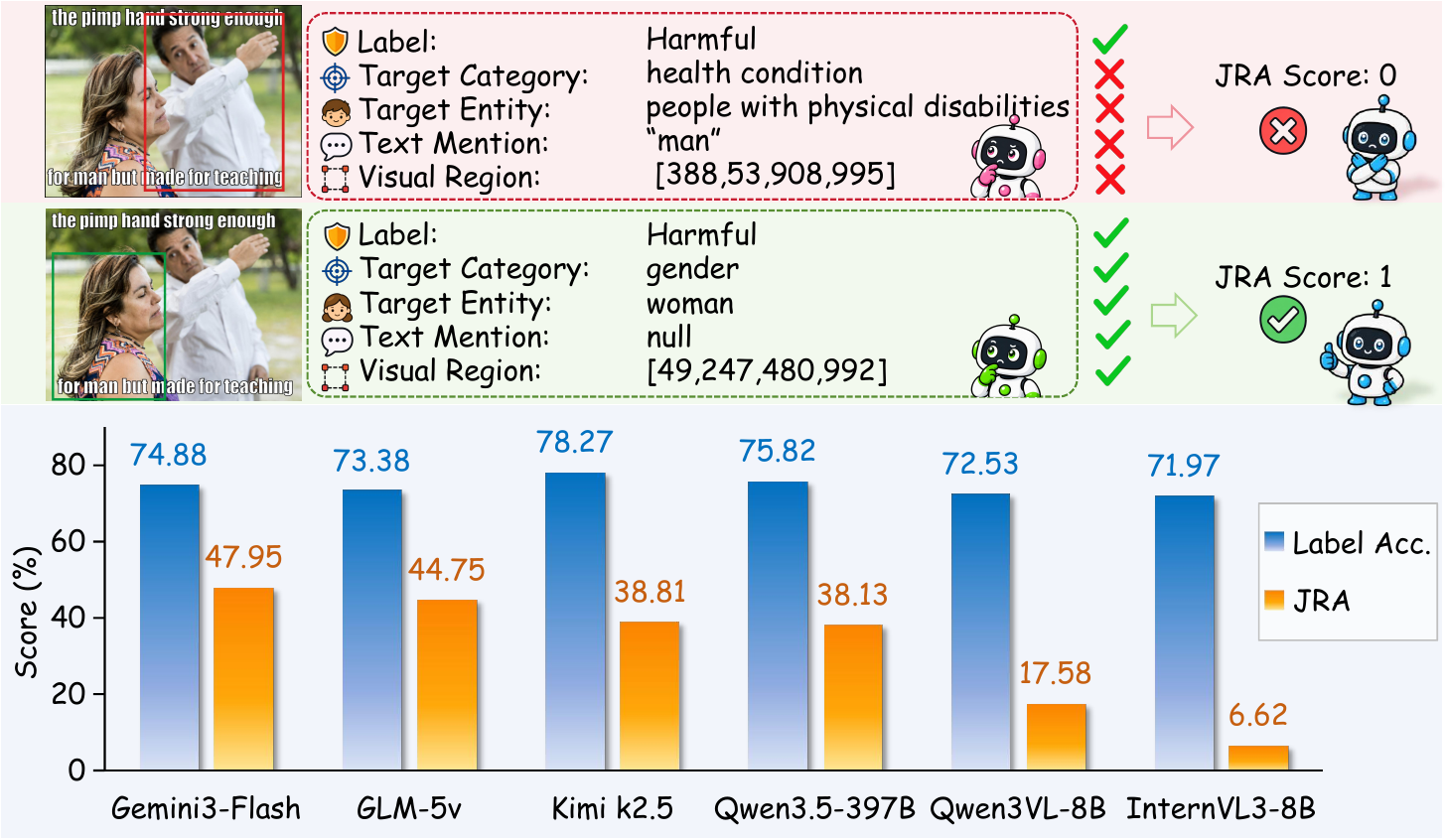}
\caption{JRA requires the harmfulness label and all target-identification fields to be jointly correct. Under this criterion, representative MLLMs show
a clear gap between harmfulness accuracy and fine-grained
target identification.}
\label{fig:structured-attribution-failure}
\end{figure}

For harmful memes, these fields form a traceable record linking the
harmfulness judgment to the attacked target and the textual and visual evidence
supporting its identification. This record allows reviewers to assess whether
the target and evidence are consistent with the final judgment. The EU Digital
Services Act emphasizes that moderation decisions should be reasoned and open
to review and contestation by affected users
\cite{item7}. Fine-grained target identification provides the
structured information needed for this examination by making the attacked
target and its multimodal grounding explicit.

The reliability of this record depends on the joint correctness of the
harmfulness label and all target-identification fields. Existing
harmful meme datasets provide annotations for only subsets of these fields
\cite{Hee2023DecodingTU,shah2024memeclip,bui-etal-2025-multi3hate}.
They therefore cannot support record-level evaluation of their joint
correctness. To address this gap, we introduce \textbf{Meme3W}, a dataset for
fine-grained target identification in harmful memes. Meme3W integrates
multiple public datasets under a unified schema and provides human-verified
annotations for harmful instances. We further introduce \textbf{Joint Record
Accuracy} (JRA), a strict record-level metric requiring the harmfulness label
and all target-identification fields to be jointly correct.
Evaluation of representative MLLMs on Meme3W reveals a clear gap between
harmfulness prediction and jointly correct fine-grained target identification.
As shown in Figure~\ref{fig:structured-attribution-failure}, harmfulness
accuracy consistently exceeds JRA across model scales. The best JRA among
general-purpose MLLMs is 47.95\%, while smaller MLLMs generally remain below
25\%.

To narrow this gap, we propose \textbf{HarmTrace}, an anchor-calibrated
decoupled optimization framework for fine-grained target identification in
harmful memes. HarmTrace first performs entity-aware supervised initialization
and then applies conditional policy optimization. Unlike standard SFT with
uniform token weighting, Entity-aware Supervised Fine-Tuning (E-SFT) upweights
target-entity tokens, strengthening supervision for the short field that
connects the target category to textual and visual grounding. Rather than using
a single aggregate reward, Conditional Target-identification Policy Optimization
(CTPO) separately normalizes harmfulness and target-identification advantages.
It also restricts target-identification updates to label-correct harmful
responses. Within CTPO, a Virtual Positive Anchor (VPA) adds a virtual reward
representing complete correctness during target-identification advantage
normalization. Experiments on two MLLM backbones show substantial improvements.
HarmTrace raises JRA from 17.58\% to 52.51\% on Qwen3-VL-8B and from 6.62\% to
49.09\% on InternVL3-8B. It also improves all evaluated target-identification
fields on both backbones.

Our contributions are as follows:
\begin{itemize}

\item We extend harmful meme detection with fine-grained target identification
and construct \textbf{Meme3W}. It augments target-category annotations with
human-verified target entities and available textual and visual grounding.
We further introduce \textbf{Joint Record Accuracy} (JRA) for strict
record-level evaluation.

\item We propose \textbf{HarmTrace}, an anchor-calibrated decoupled optimization
framework. It strengthens target-entity supervision, decouples harmfulness and
target-identification credit assignment, and calibrates target-identification
advantages with VPA.

\item We evaluate representative MLLMs on Meme3W, reveal a clear gap between
harmfulness prediction and fine-grained target identification, and show that
HarmTrace improves JRA across the evaluated backbones.

\end{itemize}

\section{Related Work}

\subsection{Multimodal Hateful Meme Detection}

Harmful meme understanding is moving beyond binary classification toward
target prediction. Hateful Memes focuses on hateful/non-hateful classification
\cite{kiela2020hateful}, while MAMI adds misogyny-type prediction
\cite{fersini2022semeval}. Harm-C and PrideMM provide harmfulness levels or
coarse target annotations \cite{pramanick2021momenta,shah2024memeclip};
MemeMind and MemeIntel add chain-of-thought or explanation-oriented
supervision \cite{Gu2025MemeMindAL,Kmainasi2025MemeIntelED}; and MemeLens
unifies multilingual and multitask meme resources \cite{memelens2025}.
However, these datasets do not provide unified annotations for the concrete
target entity and its supporting textual and visual evidence, nor formulate
these fields jointly with harmfulness prediction. Recent work improves harmful meme detection through multimodal representation
learning \cite{burbi2023mapping} and retrieval- or knowledge-augmented
adaptation \cite{mei-etal-2024-improving,Tzelepi2025ImprovingMH,
Mei2025RobustAO}. Other methods introduce multimodal debate or generated
explanations \cite{Lin2024TowardsEH,Hee2025DemystifyingHC,
mei2026expohmlearningexplainthendetecthateful}, while DR-HM adopts reasoning-enhanced training
\cite{Cheng2026DRHMDT}. These methods primarily optimize binary harmfulness
classification. Fine-grained target identification is not included in their
prediction objectives.

\subsection{Reinforcement Learning for MLLMs}

Reinforcement learning has become an important post-training paradigm for
improving the reasoning and alignment of LLMs and MLLMs. PPO stabilizes policy
updates through clipping and a learned value function
\cite{schulman2017proximal}, DPO directly optimizes preference pairs
\cite{rafailov2023direct}, and GRPO estimates advantages within sampled groups
without a critic \cite{shao2024deepseekmath}. Recent variants further refine
group-based optimization through dynamic sampling, process-level rewards, and
negative-enhanced signals for all-negative groups
\cite{yu2025dapo,tan2026papostabilizingrubricintegration,nan2025ngrpo}. These methods are mainly studied
in mathematical, coding, and general reasoning tasks. In harmful meme
detection, HarmTrace applies group-based optimization to fine-grained target
identification.

\section{\emph{Meme3W}: A Fine-Grained Target Identification Dataset for Harmful Memes}
\subsection{Task Definition and Output Schema}

We extend multimodal harmful meme detection with fine-grained target
identification. The task predicts harmfulness for every meme and, for harmful
memes, additionally identifies the target category, target entity, textual
mention, and visual region. This extension enables evaluation beyond label correctness by
making the attacked target and its available textual and visual evidence explicit.
Given a dataset
$\mathcal{D}=\{(x_i,a_i^\star)\}_{i=1}^{N}$, each input
$x_i=(I_i,T_i)$ consists of a meme image $I_i$ and its associated text $T_i$.
The structured annotation is defined as
\begin{equation}
    a_i^\star
    =
    (l_i^\star,c_i^\star,e_i^\star,s_i^\star,B_i^\star),
    \label{eq:structured_answer}
\end{equation}
where $l_i^\star\in\{\mathrm{harmful},\mathrm{non\mbox{-}harmful}\}$ denotes
the harmfulness label, $c_i^\star$ denotes the category of the attacked target,
$e_i^\star$ denotes the specific target entity, $s_i^\star$ denotes the
textual target mention, and $B_i^\star$ denotes a list of visual target
regions represented by bounding boxes.

\begin{figure*}[t]
\centering
\includegraphics[width=\textwidth]{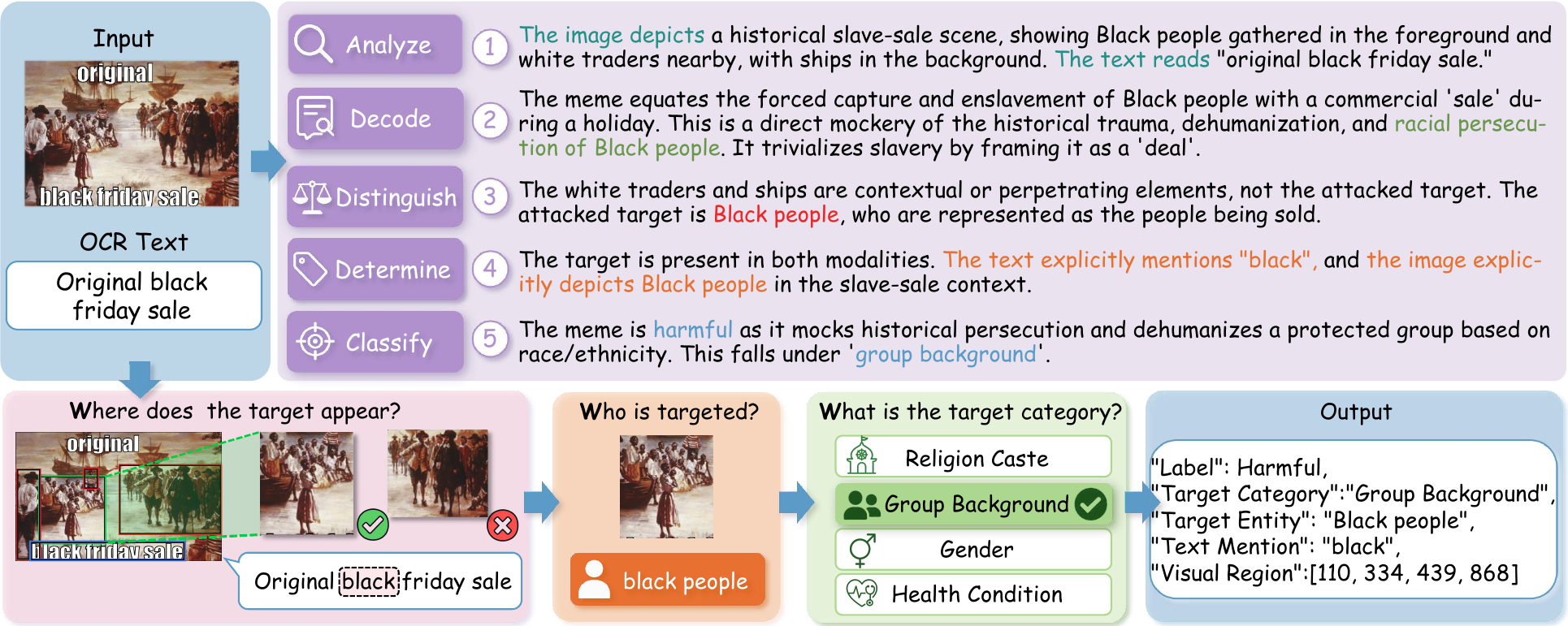}
\caption{A representative Meme3W annotation, showing the analysis process
and a final structured output for fine-grained target identification,
answering where the target appears, who is targeted, and which target
category applies.}
\label{fig:dataset_annotation}
\end{figure*}

For harmful samples, the category of the attacked target and the target entity
are annotated, while the textual and visual grounding fields are populated
only when the corresponding evidence is present. For non-harmful samples,
there is no attacked target to identify. Therefore, all non-box target fields
are set to \texttt{null}, while the visual-region field is set to
\texttt{[]}.

\subsection{Dataset Construction}

We construct Meme3W by curating samples from four public datasets:
PrideMM~\cite{shah2024memeclip}, MAMI~\cite{fersini2022semeval}, Hateful
Memes~\cite{kiela2020hateful}, and Harm-C~\cite{pramanick2021momenta}.
Their original annotations do not jointly identify the attacked target and its
textual and visual grounding. We therefore re-annotate the selected samples
under the unified schema in Eq.~\ref{eq:structured_answer}.

Before annotation, we manually screened the collected samples and removed those
whose visual content was unclear or could not be reliably recognized. The resulting
benchmark contains 10,662 samples, including 4,418 harmful samples. Harmful
samples are organized into four broad target categories: group background,
gender, religion and caste, and health condition. Before splitting, we removed
exact duplicates and assigned verified perceptual near-duplicate groups to the
same split. Meme3W is partitioned into
training (85\%), validation (5\%), and test (10\%) sets.

\subsection{Annotation Pipeline}

We construct the structured annotations through a three-stage
MLLM-assisted human annotation pipeline. Figure~\ref{fig:dataset_annotation}
presents a representative Meme3W annotation, including the analysis process
and the resulting fine-grained target-identification output.

\paragraph{Stage 1: MLLM Candidate Initialization.}
We employ Gemini-3-Flash~\cite{doshi2025gemini},
GPT-5.2~\cite{openai2025gpt52}, and
Qwen3.5-397B~\cite{qwen3.5} to independently generate one candidate
annotation for each sample. All three models follow the same annotation
prompt, which specifies five analysis steps and requires a final structured
annotation. The outputs are normalized to the predefined format, and malformed
fields are removed. The model identities are hidden, and the three candidates
are randomly ordered for each sample to reduce model-specific anchoring.

\paragraph{Stage 2: Human Expert Annotation.}
Five trained graduate annotators with strong English proficiency and NLP
backgrounds participate in this stage. They are trained on the task definition,
output schema, and annotation guidelines. Each sample is independently reviewed
by two annotators. The MLLM outputs serve only as editable references, and
annotators may revise any field based on the sample evidence. Before
adjudication, the field-specific agreement scores are Cohen's $\kappa=0.908$
for target category, normalized agreement of 0.813 for target entity, token-F1
of 0.824 for textual mention, and IoU@0.75 of 0.818 for visual regions. 
For target entities, semantically equivalent mentions are mapped to a shared canonical form before agreement computation.

\paragraph{Stage 3: Human Expert Adjudication.}
Disagreements are independently reviewed by a third annotator and resolved
through discussion among all three annotators. We further conduct a
candidate-blind audit on 200 sampled cases, in which annotators receive only
the raw samples and annotation guideline. Their annotations achieve an
unweighted mean of 0.827 across the four field-specific agreement scores
against the final gold annotations.

\begin{figure*}[t]
\centering
\includegraphics[width=\textwidth]{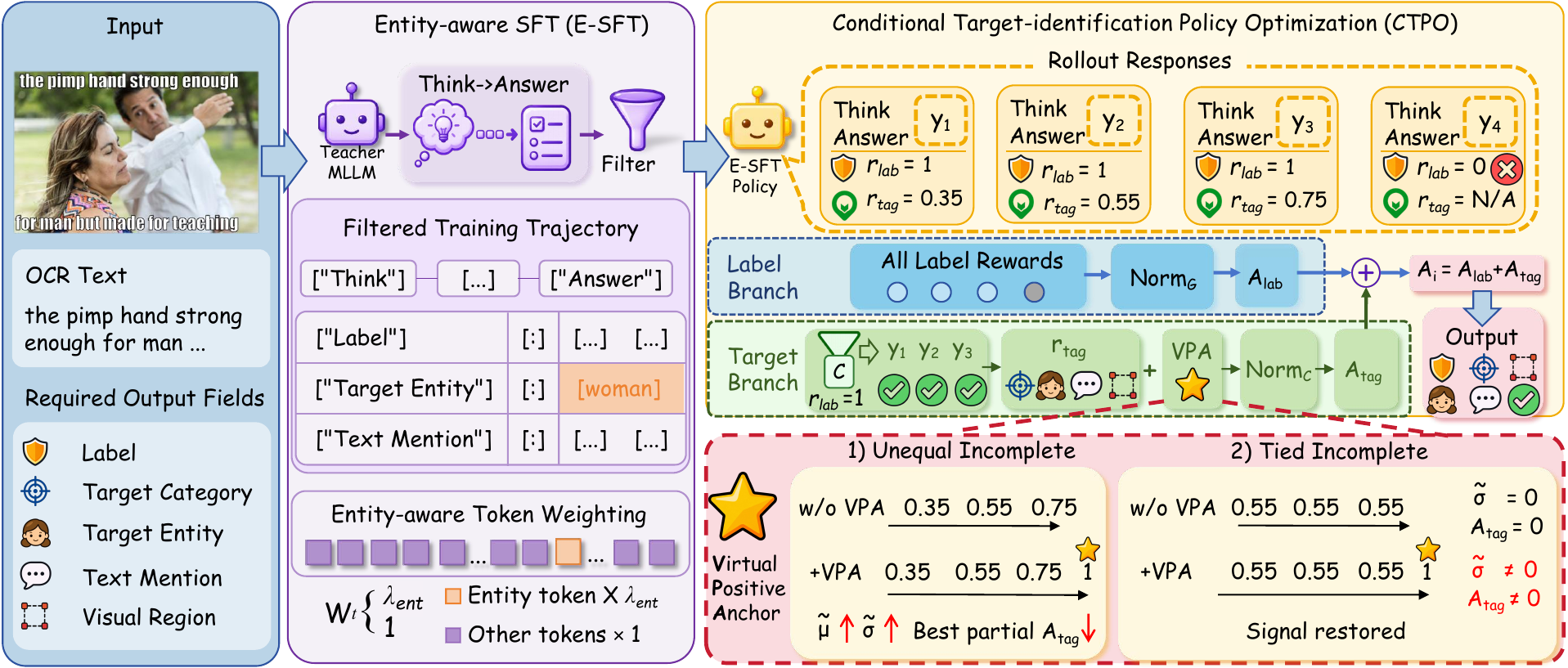}
\caption{Overview of HarmTrace. The left panel shows an input and its required output schema. E-SFT upweights target-entity tokens, while CTPO normalizes label rewards over all rollouts and target-identification rewards over the label-correct harmful subset $\mathcal{C}$. The bottom panel shows VPA for unequal and tied rewards among incomplete responses.}
\label{fig:framework}
\end{figure*}

\section{Method}

\subsection{Framework Overview}
\label{sec:HarmTrace_overview}

To improve the joint correctness of harmfulness and target-identification
predictions, we propose \textbf{HarmTrace}, an anchor-calibrated decoupled
optimization framework. Figure~\ref{fig:framework} provides an overview.
HarmTrace consists of Entity-aware Supervised Fine-Tuning (E-SFT) and
Conditional Target-identification Policy Optimization (CTPO). E-SFT upweights
target-entity tokens to strengthen entity supervision. CTPO decouples
harmfulness and target-identification advantages and applies
target-identification optimization only to label-correct harmful responses.
Within CTPO, a Virtual Positive Anchor (VPA) augments target-identification
advantage normalization with a virtual fully correct response.
For training, HarmTrace adopts an explain-then-answer sequence:
\begin{equation}
    y_i
    =
    \{\mathrm{think}:z_i,\ \mathrm{answer}:a_i\},
    \label{eq:training_sequence}
\end{equation}
where the answer field follows Eq.~(\ref{eq:structured_answer}), and the
think field serves as an intermediate reasoning scaffold for generating the
structured answer.

\subsection{Entity-aware Supervised Fine-Tuning (E-SFT)}
\label{sec:esft}

Before policy optimization, we initialize the policy through cold-start
supervised fine-tuning. We use Gemini-3-Flash to generate one
explain-then-answer trajectory for each training meme $x_i$. Both trajectory
generation and filtering use only the training split. We retain only
trajectories whose structured answers are valid and consistent with the
verified gold annotations:
\begin{equation}
    \mathcal{D}_{\mathrm{cold}}
    =
    \left\{
        (x_i,y_i)
        \;\middle|\;
        \mathrm{Match}(y_i,a_i^\star)=1
    \right\},
    \label{eq:cold_dataset}
\end{equation}
where $\mathrm{Match}(\cdot)$ extracts the structured answer from $y_i$ and
checks its format and consistency with $a_i^\star$. The filtered trajectories
are used for cold-start training, whereas CTPO uses the full training set.
Standard SFT and E-SFT share the same trajectories and differ only in token
weighting. The target entity identifies who is attacked and links the target
category to textual and visual evidence, yet its short value contributes little
to the sequence loss under uniform weighting. E-SFT therefore upweights the
\texttt{target\_entity} value tokens. For
$(x_i,y_i)\in\mathcal{D}_{\mathrm{cold}}$, we optimize
\begin{equation}
    \mathcal{L}_{\mathrm{E\mbox{-}SFT}}
    =
    -\frac{
        \sum_{t=1}^{T}
        w_t\log p_{\theta}
        (y_{i,t}\mid x_i,y_{i,<t})
    }{
        \sum_{t=1}^{T}w_t
    },
    \label{eq:esft_objective}
\end{equation}
where
\begin{equation}
    w_t
    =
    \left\{
    \begin{array}{ll}
        \lambda_{\mathrm{ent}},
        & t\in\mathcal{I}_{\mathrm{ent}},\\
        1,
        & \mbox{otherwise},
    \end{array}
    \right.
    \label{eq:entity_weight}
\end{equation}
$\mathcal{I}_{\mathrm{ent}}$ denotes the target-entity value positions and
$\lambda_{\mathrm{ent}}>1$ controls their supervision weight. The resulting
policy initializes CTPO.

\subsection{Conditional Target-identification Policy Optimization (CTPO)}
\label{sec:ctpo}

For each input meme $x$, the old policy
$\pi_{\theta_{\mathrm{old}}}$ samples a group of responses
$\{y_i\}_{i=1}^{G}$. Responses violating the output format or
label-conditioned schema receive zero rewards. For valid responses,
$r_i^{\mathrm{lab}}\in\{0,1\}$ measures harmfulness-label correctness, while
$r_i^{\mathrm{tag}}\in[0,1]$ combines applicable field scores with
schema-applicability consistency. The latter penalizes missing required
grounding and hallucinated non-applicable grounding, and is computed only for
label-correct harmful responses, with $r_i^{\mathrm{tag}}=1$ indicating
complete correctness. Detailed reward weights and matching functions are
provided in Appendix B.3.

Combining the two rewards would mix label and target-identification credit.
CTPO therefore computes their advantages separately. The label advantage is
normalized over the full rollout group:

\begin{equation}
A_i^{\mathrm{lab}}
=
\frac{
r_i^{\mathrm{lab}}-\mu_{\mathrm{lab}}
}{
\sigma_{\mathrm{lab}}+\epsilon_{\mathrm{norm}}
},
    \label{eq:label_advantage}
\end{equation}
where $\mu_{\mathrm{lab}}$ and $\sigma_{\mathrm{lab}}$ are computed from
$\{r_j^{\mathrm{lab}}\}_{j=1}^{G}$, and
$\epsilon_{\mathrm{norm}}>0$ is a small constant for numerical stability.

Target-identification learning is restricted to
\begin{equation}
\mathcal{C}
=
\{j\mid r_j^{\mathrm{lab}}=1,\;
l^\star=\mathrm{harmful}\},
    \label{eq:correct_subset}
\end{equation}
and responses outside $\mathcal{C}$ receive zero target-identification
advantage.

Normalizing target-identification rewards only within $\mathcal{C}$ has two
limitations when all responses in $\mathcal{C}$ are incomplete. With unequal
rewards, the best partial response is normalized only against other incomplete
responses and can therefore receive a strong positive advantage. With tied
rewards, the reward variance becomes zero and all target-identification
advantages vanish. We therefore introduce a \emph{Virtual Positive Anchor}
(VPA) by adding a virtual fully correct score $r_{\max}=1$ to the normalization
multiset:
\begin{equation}
    \widetilde{\mathcal{R}}_{\mathcal{C}}^{\mathrm{tag}}
    =
    \{r_j^{\mathrm{tag}}\mid j\in\mathcal{C}\}
    \uplus\{r_{\max}\}.
    \label{eq:vpa_set}
\end{equation}
When $\mathcal{C}$ is not empty, VPA is included only in the normalization
statistics and contributes no policy-loss term. Let $m=|\mathcal{C}|$. The
resulting statistics are
\begin{equation}
\begin{array}{@{}rl@{}}
    \widetilde{\mu}_{\mathcal{C}}^{\mathrm{tag}}
    &=
    \displaystyle
    \frac{
        \sum_{j\in\mathcal{C}}r_j^{\mathrm{tag}}+r_{\max}
    }{
        m+1
    },\\[0.8em]
    \widetilde{\sigma}_{\mathcal{C}}^{\mathrm{tag}}
    &=
    \displaystyle
    \sqrt{
        \frac{1}{m+1}
        \sum_{r\in
        \widetilde{\mathcal{R}}_{\mathcal{C}}^{\mathrm{tag}}}
        \left(
            r-\widetilde{\mu}_{\mathcal{C}}^{\mathrm{tag}}
        \right)^2
    }.
\end{array}
\label{eq:vpa_statistics}
\end{equation}
The target-identification advantage is
\begin{equation}
    A_i^{\mathrm{tag}}
    =
    \left\{
    \begin{array}{ll}
        \displaystyle
        \frac{
            r_i^{\mathrm{tag}}
            -
            \widetilde{\mu}_{\mathcal{C}}^{\mathrm{tag}}
        }{
            \widetilde{\sigma}_{\mathcal{C}}^{\mathrm{tag}}
            +
            \epsilon_{\mathrm{norm}}
        },
        & i\in\mathcal{C},\\[1.0em]
        0,
        & i\notin\mathcal{C}.
    \end{array}
    \right.
    \label{eq:tag_advantage}
\end{equation}
If $\mathcal{C}$ is empty, we set $A_i^{\mathrm{tag}}=0$ for all responses.
The final advantage combines the separately normalized terms:
\begin{equation}
    A_i
    =
    A_i^{\mathrm{lab}}
    +
    A_i^{\mathrm{tag}}.
    \label{eq:total_advantage}
\end{equation}

CTPO uses $A_i$ in the standard clipped GRPO objective, with the policy
obtained after E-SFT fixed as $\pi_{\mathrm{ref}}$. Let
$\rho_{i,t}(\theta)$ denote the token-level probability ratio between
$\pi_{\theta}$ and $\pi_{\theta_{\mathrm{old}}}$. The optimization objective is
\begin{equation}
\begin{array}{@{}l@{}}
    \mathcal{J}_{\mathrm{CTPO}}(\theta)
    =
    \mathrm{E}_{x,\{y_i\}\sim\pi_{\theta_{\mathrm{old}}}}
    \bigl[
    \displaystyle
    \frac{1}{G}\sum_{i=1}^{G}
    \frac{1}{|y_i|}\sum_{t=1}^{|y_i|}
    \\[0.2em]
    \quad
    \displaystyle
    \bigl(\min\!\bigl(
    \rho_{i,t}(\theta)A_i,\,
    \mathrm{clip}\!\bigl(
    \rho_{i,t}(\theta),
    1-\epsilon_{\mathrm{clip}},
    1+\epsilon_{\mathrm{clip}}
    \bigr)A_i
    \bigr)
    \\[0.2em]
    \quad
    \displaystyle
    {}-\beta
    D_{\mathrm{KL}}\!\bigl(
    \pi_{\theta}
    \,\|\,
    \pi_{\mathrm{ref}}
    \bigr)
    \bigr)
    \bigr].
\end{array}
    \label{eq:ctpo_objective}
\end{equation}

\section{Experiments}

\subsection{Experimental Setup}

\paragraph{Baselines.}
We evaluate HarmTrace on Meme3W against three baseline groups.
1) \textbf{General-purpose MLLMs}, including
Gemini 3 Flash~\cite{doshi2025gemini},
GPT-5.2~\cite{openai2025gpt52},
GLM-5V-Turbo~\cite{glm5v},
Qwen3-VL models at multiple parameter scales~\cite{Bai2025Qwen3VLTR},
Qwen3.5/3.6~\cite{qwen3.5,qwen3.6-27b},
InternVL3/3.5~\cite{item12,Wang2025InternVL35AO},
Gemma~\cite{team2024gemma}, and
Kimi-K2.5~\cite{Bai2026KimiKV}.
Table~\ref{tab:main-results} lists all models.
2) \textbf{Harmful meme detection methods}.
We adapt EXPO-HM~\cite{mei2026expohmlearningexplainthendetecthateful} from our standard SFT checkpoint and
apply its RL optimization method to the fine-grained target-identification task.
3) \textbf{RL-based optimization methods}.
Under the same E-SFT initialization, we compare HarmTrace with
PPO~\cite{schulman2017proximal},
GRPO~\cite{shao2024deepseekmath},
DAPO~\cite{yu2025dapo}, and
PAPO~\cite{tan2026papostabilizingrubricintegration}.

\begin{table*}[!t]
\centering
\begin{tabular*}{\textwidth}{@{\extracolsep{\fill}}lccccccccc@{}}
\toprule
\multirow{2}{*}{\textbf{Model}} &
\multicolumn{2}{c}{\textbf{Label}} &
\multirow{2}{*}{\textbf{JRA}} &
\multicolumn{1}{c}{\textbf{Category}} &
\multicolumn{1}{c}{\textbf{Entity}} &
\multicolumn{2}{c}{\textbf{Text Mention}} &
\multicolumn{2}{c}{\textbf{Visual Region}} \\
\cmidrule(lr){2-3}
\cmidrule{5-5}
\cmidrule{6-6}
\cmidrule(lr){7-8}
\cmidrule(lr){9-10}
& Acc. & F1 & & EM & $\mathrm{F1}_{\mathrm{rel}}$ & EM & F1 & IoU50 & IoU75 \\
\midrule
\multicolumn{10}{c}{\textit{Closed-source MLLMs}} \\
\cmidrule(lr){1-10}
Gemini3-Flash & 74.88 & 73.66 & 47.95 & \textbf{79.91} & \textbf{73.29} & \textbf{61.87} & 63.01 & 65.98 & 59.59 \\
GPT-5.2 & 75.92 & 69.01 & 28.31 & 62.56 & 60.73 & 43.38 & 46.80 & 38.58 & 33.33 \\
GLM-5V-Turbo & 73.38 & 72.28 & 44.75 & 74.89 & 68.26 & \underline{60.96} & 62.56 & 62.79 & 57.31 \\
\midrule
\multicolumn{10}{c}{\textit{Open-source Large MLLMs}} \\
\cmidrule(lr){1-10}
Kimi-K2.5 & 78.27 & 75.86 & 38.81 & \underline{76.26} & \underline{73.06} & 55.48 & 56.85 & 59.59 & 55.02 \\
Gemma-4-26B-A4B & 78.46 & 74.24 & 28.54 & 67.81 & 57.08 & 52.97 & 54.11 & 62.79 & 41.78 \\
Qwen3.5-397B-A17B & 75.82 & 74.58 & 38.13 & 75.11 & 70.78 & 59.36 & \underline{63.70} & \textbf{73.06} & 59.13 \\
Qwen3-VL-235B-A22B & 76.01 & 70.99 & 21.46 & 63.47 & 52.97 & 37.44 & 39.50 & 56.39 & 47.49 \\
Qwen3.6-27B & 76.95 & 73.80 & 36.99 & 69.63 & 61.87 & 53.42 & 55.25 & 63.24 & \underline{59.82} \\
Qwen3.5-27B & 75.16 & 73.65 & 35.84 & 74.89 & 65.98 & 52.05 & 53.65 & 63.93 & 59.36 \\
Qwen3-VL-32B & 75.26 & 73.51 & 29.00 & 76.03 & 67.35 & 50.91 & 52.05 & 65.98 & 50.68 \\
\midrule
\multicolumn{10}{c}{\textit{Open-source Small MLLMs}} \\
\cmidrule(lr){1-10}
EXPO-HM (Qwen3-VL-8B) & 78.93 & 74.58 & 48.86 & 70.55 & 65.53 & 57.76 & 59.36 & 62.10 & 58.68 \\
Qwen3.5-9B & 75.54 & 68.22 & 23.06 & 53.65 & 45.43 & 41.10 & 42.47 & 52.51 & 42.92 \\
Qwen3-VL-8B & 73.75 & 69.03 & 17.58 & 60.05 & 46.35 & 34.02 & 37.67 & 53.20 & 39.95 \\
InternVL3.5-8B & 68.77 & 67.95 & 14.61 & 64.84 & 44.06 & 36.76 & 38.36 & 56.85 & 34.93 \\
InternVL3-8B & 71.97 & 68.96 & 6.62 & 60.05 & 46.35 & 17.81 & 19.63 & 53.20 & 36.53 \\
Qwen3-VL-4B & 72.53 & 66.20 & 10.96 & 57.99 & 37.90 & 24.89 & 27.40 & 50.00 & 31.05 \\
\midrule
\multicolumn{10}{c}{\textit{Ours}} \\
\cmidrule(lr){1-10}
HarmTrace (InternVL3-8B) & \underline{80.06} & \underline{76.02} & \underline{49.09} & 71.23 & 66.21 & 58.22 & 62.84 & 60.73 & 56.85 \\
HarmTrace (Qwen3-VL-8B) & \textbf{80.15} & \textbf{76.77} & \textbf{52.51} & 72.60 & 68.72 & 60.73 & \textbf{64.93} & \underline{66.67} & \textbf{62.56} \\
\bottomrule
\end{tabular*}
\caption{Main results of different models on Meme3W. Best results are bolded, and second-best results are underlined.}
\label{tab:main-results}
\end{table*}

\begin{table*}[!t]
\centering
\begin{tabular*}{\textwidth}{@{\extracolsep{\fill}}lccccccccc@{}}
\toprule
\multirow{2}{*}{\textbf{Setting}} &
\multicolumn{2}{c}{\textbf{Label}} &
\multirow{2}{*}{\textbf{JRA}} &
\multicolumn{1}{c}{\textbf{Category}} &
\multicolumn{1}{c}{\textbf{Entity}} &
\multicolumn{2}{c}{\textbf{Text Mention}} &
\multicolumn{2}{c}{\textbf{Visual Region}} \\
\cmidrule(lr){2-3}
\cmidrule{5-5}
\cmidrule{6-6}
\cmidrule(lr){7-8}
\cmidrule(lr){9-10}
& Acc. & F1 & & EM & $\mathrm{F1}_{\mathrm{rel}}$ & EM & F1 & IoU50 & IoU75 \\
\midrule
Qwen3-VL-8B & 73.75 & 69.03 & 17.58 & 60.05 & 46.35 & 34.02 & 37.67 & 53.20 & 39.95 \\
+SFT & \underline{77.80} & \underline{73.00} & 43.15 & 65.98 & 61.42 & 52.05 & 56.42 & 57.53 & 53.88 \\
+E-SFT ($\lambda_{\mathrm{ent}}=5$) & 77.33 & 72.95 & 44.75 & \textbf{67.81} & 62.79 & \textbf{55.94} & \textbf{59.92} & \underline{58.90} & \underline{54.34} \\
+E-SFT ($\lambda_{\mathrm{ent}}=10$) & \textbf{78.65} & \textbf{74.26} & \textbf{45.89} & \underline{67.58} & \underline{63.70} & 54.79 & \underline{58.67} & \textbf{61.19} & \textbf{57.31} \\
+E-SFT ($\lambda_{\mathrm{ent}}=15$) & 77.14 & 72.35 & \underline{44.98} & 66.67 & \textbf{64.16} & \underline{55.02} & \underline{58.67} & 57.53 & 54.11 \\
\bottomrule
\end{tabular*}
\caption{Ablation study of SFT and E-SFT with different entity-token weights $
  \lambda_{\mathrm{ent}}$.}
\label{tab:sft-ablation}
\end{table*}

\paragraph{Metrics.}
\label{sec:jra}
We report three groups of metrics. 
1) \textbf{Harmfulness detection.}
We evaluate binary harmfulness classification using accuracy and F1 over all
test samples. 
2) \textbf{Joint record correctness.} Joint Record Accuracy (JRA) is computed over all gold-harmful samples. 
A record is correct only when the meme is predicted as harmful and every
target-identification field satisfies its respective matching criterion. For a field that is
absent in the gold record, the prediction must also indicate absence. Let
$\mathcal{H}=\{i\mid l_i^\star=\mathrm{harmful}\}$ denote the set of
gold-harmful samples and $\mathcal{F}=\{c,e,s,B\}$ the
target-identification fields. Here, $\hat l_i$ and $l_i^\star$ denote the
predicted and gold harmfulness labels. For each field $f$, $\hat f_i$,
$f_i^\star$, $S_f$, and $\tau_f$ denote its predicted value, gold value,
matching score, and correctness threshold, respectively. We define
\begin{equation}
    \mathrm{JRA}
    =
    \frac{1}{|\mathcal{H}|}
    \sum_{i\in\mathcal{H}}
    \mathbf{1}
    \left[
        \hat l_i=l_i^\star
    \right]
    \prod_{f\in\mathcal{F}}
    \mathbf{1}
    \left[
        S_f(\hat f_i,f_i^\star)\geq\tau_f
    \right],
    \label{eq:jra}
\end{equation}
where
$(S_c,S_e,S_s,S_B)
=(\mathrm{EM},\mathrm{F1}_{\mathrm{rel}},
\mathrm{F1}_{\mathrm{tok}},\mathrm{IoU})$
and
$(\tau_c,\tau_e,\tau_s,\tau_B)
=(1,0.7,0.7,0.5)$.
Semantically equivalent target-entity mentions are canonicalized before
computing $\mathrm{F1}_{\mathrm{rel}}$. Following prior token-overlap-based
relaxed evaluation~\citep{heo-etal-2025-large}, we set
$\tau_e=\tau_s=0.7$ to allow minor lexical differences in target entities and
boundary variations in textual mentions. For visual regions, IoU is computed
between the minimum enclosing rectangles of all predicted and gold boxes
\cite{yu2016unitbox}.
3) \textbf{Field-level diagnosis.}
On gold-harmful samples, we report EM for target category,
$\mathrm{F1}_{\mathrm{rel}}$ for target entity, EM and token-F1 for textual
mention, and set accuracy for visual regions at IoU thresholds of $0.5$ and
$0.75$.

\paragraph{Settings.}
All trainable methods were run for 3 epochs on 2$\times$NVIDIA H200 GPUs using
LoRA \cite{Hu2021LoRALA} with rank $=64$ and $\alpha=128$. RL training uses a rollout group size
of 8. Additional SFT and RL hyperparameters are provided in Appendix~B.1.

\subsection{Main Results}

Table~\ref{tab:main-results} shows that current MLLMs achieve higher
harmfulness accuracy than JRA, revealing a clear gap between harmfulness
detection and jointly correct fine-grained target identification. Among the
general-purpose MLLMs, the best JRA is 47.95\%, while most smaller
general-purpose MLLMs remain below 25\% despite substantially higher
harmfulness accuracy. Strong performance on individual fields also does not
necessarily translate into a jointly correct record. The absence of a
consistent advantage for larger models further suggests that scaling alone
does not resolve the difficulty of producing jointly correct
target-identification outputs. HarmTrace narrows this gap on both evaluated
backbones, raising JRA from 17.58\% to 52.51\% on Qwen3-VL-8B and from
6.62\% to 49.09\% on InternVL3-8B, corresponding to absolute improvements of
34.93 and 42.47 points. Harmfulness accuracy and F1 also improve on both
backbones, indicating that the gains in target identification do not come at
the expense of harmfulness detection. On Qwen3-VL-8B, HarmTrace also improves
all evaluated target-identification fields over the base model. These gains
are consistent with the design of HarmTrace, which strengthens fine-grained
target identification while maintaining harmfulness detection through
decoupled credit assignment. Overall, HarmTrace improves both field-level
target identification and the joint correctness of harmfulness and
target-identification outputs.

\subsection{Ablation and Mechanism Analysis}

To understand how each component contributes to HarmTrace, we conduct targeted
ablations on E-SFT, decoupled optimization, and the VPA.

\paragraph{Entity-Aware Supervision Weight.}
Since the target entity links harmfulness detection to supporting textual and
visual evidence, we examine whether stronger entity supervision improves the
supervised initialization. Using the same explain-then-answer trajectories,
Table~\ref{tab:sft-ablation} compares standard SFT with E-SFT variants using
different entity-token weights. Standard SFT raises JRA from 17.58\% to 43.15\%, while
all E-SFT variants provide further gains. Among the tested weights,
$\lambda_{\mathrm{ent}}=10$ provides the most balanced performance, with a JRA
of 45.89\% and strong results on harmfulness and visual-region metrics.
Although the other weights perform better on a few individual fields, their
lower JRA suggests less consistent performance across the full output. Overall,
E-SFT consistently improves over standard SFT across the tested weights,
with $\lambda_{\mathrm{ent}}=10$ showing a favorable balance between JRA,
harmfulness detection, and visual-region performance.

\paragraph{RL-Stage Components.}

\begin{table}[!t]
\centering
\begin{tabular*}{\columnwidth}{@{\extracolsep{\fill}}l*{4}{c}@{}}
\toprule
\multirow{2}{*}{\textbf{Init.}} &
\multirow{2}{*}{\textbf{RL}} &
\multicolumn{2}{c}{\textbf{Label}} &
\multirow{2}{*}{\textbf{JRA}} \\
\cmidrule(lr){3-4}
& & Acc. & F1 & \\
\midrule
w/o SFT & GRPO & 74.51 & 71.52 & 18.72 \\
 & +Dec. & 74.41 & 71.43 & 20.32 \\
 & +VPA & 75.16 & 71.80 & 22.37 \\
\midrule
+SFT & GRPO & 77.99 & 74.89 & 45.21 \\
 & +Dec. & 77.42 & 73.74 & 47.26 \\
 & +VPA & 78.65 & 74.35 & 49.54 \\
\midrule
+E-SFT & GRPO & 78.74 & 74.49 & 49.09 \\
 & +Dec. & \underline{79.12} & \underline{75.33} & \underline{50.91} \\
 & +VPA & \textbf{80.15} & \textbf{76.77} & \textbf{52.51} \\
\bottomrule
\end{tabular*}
\caption{RL-stage ablation of HarmTrace under different starting points.
w/o SFT denotes initializing RL from the base model; +Dec. adds decoupled
reward optimization; and +VPA further adds VPA to +Dec.}
\label{tab:rl-ablation}
\end{table}

\begin{figure}[t]
    \centering
    \includegraphics[width=\columnwidth]
    {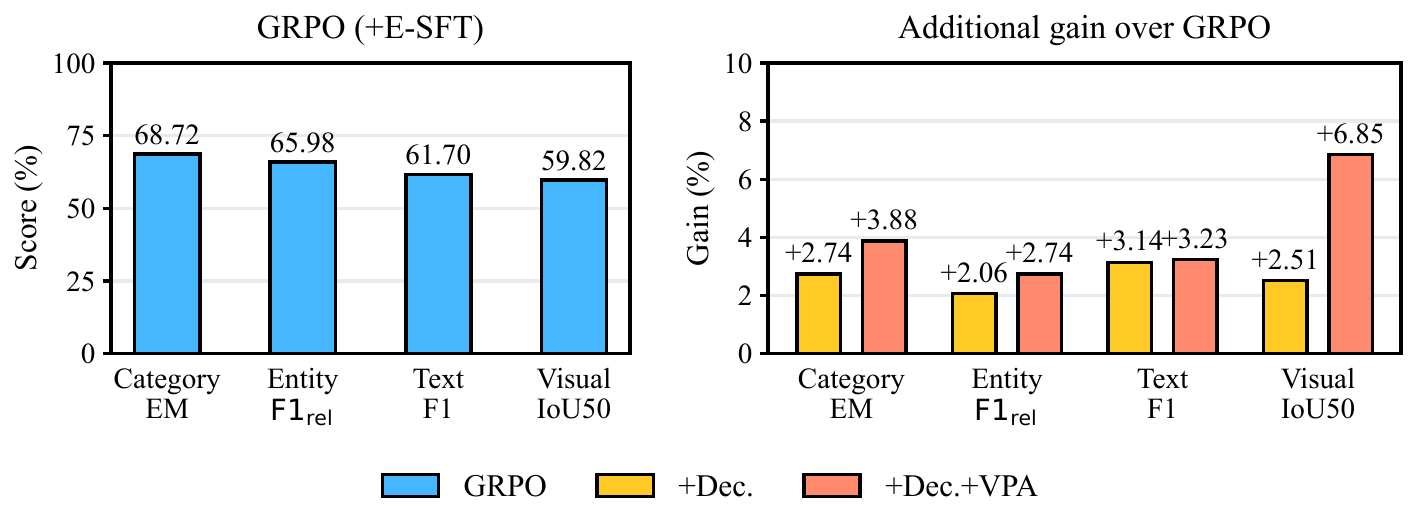}
    \caption{Target-identification results under E-SFT initialization.
    The left panel shows GRPO scores, and the right shows gains from decoupled
    optimization and VPA over GRPO.}
    \label{fig:field-ablation}
\end{figure}

\begin{figure}[!t]
    \centering
    \includegraphics[width=\columnwidth]
    {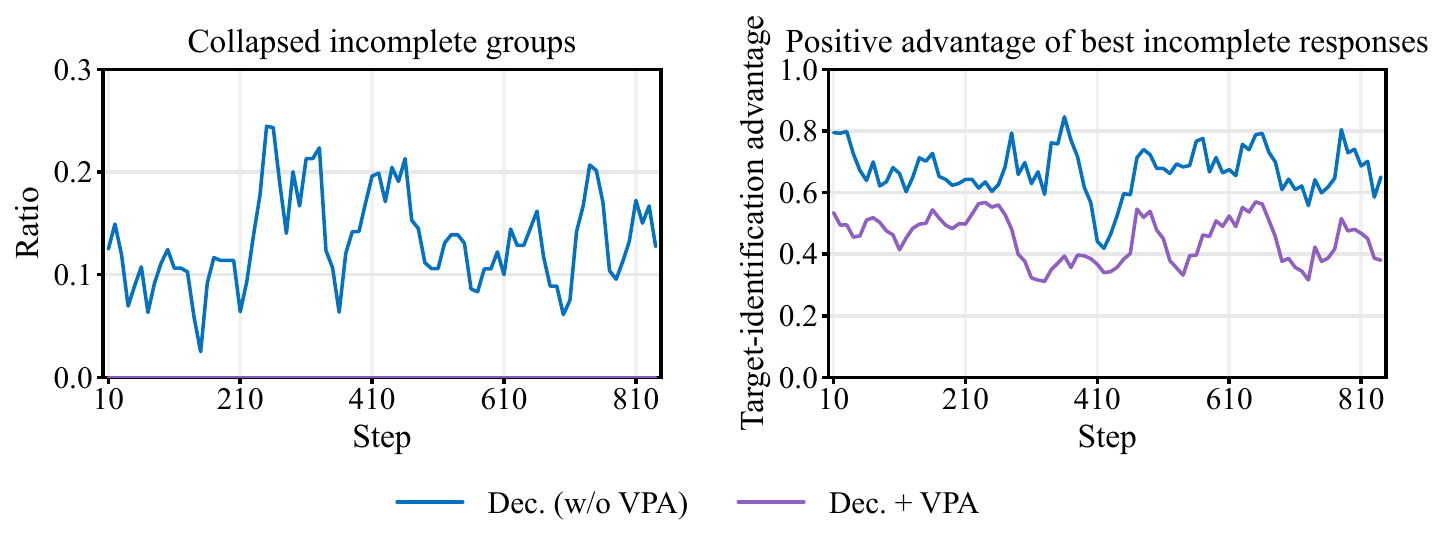}
    \caption{Effect of VPA under E-SFT initialization. The left panel shows the
    zero-advantage group ratio, while the right tracks positive advantage of
    the best incomplete responses.}
    \label{fig:vpa-mechanism}
\end{figure}

To examine whether decoupled optimization and VPA remain effective under
different supervised initializations, Table~\ref{tab:rl-ablation} evaluates
their contributions under three starting points. Under E-SFT initialization,
decoupled optimization raises JRA from 49.09\% to 50.91\%, and adding VPA
further raises it to 52.51\%. The same stepwise improvement is observed under
the other two initializations. Figure~\ref{fig:field-ablation} further shows
that, under E-SFT initialization, both components improve all evaluated
target-identification fields, indicating that the JRA improvement is not
driven by a single field. To further examine the role of VPA,
Figure~\ref{fig:vpa-mechanism} analyzes its effect on credit assignment. VPA
reduces zero-advantage incomplete groups and lowers the positive advantage of
the best incomplete responses. Overall, decoupled optimization and VPA
consistently improve JRA across initialization settings, and their gains under
E-SFT extend across all evaluated target-identification fields.

\subsection{Comparison with RL Baselines}

To determine whether the gains arise from the optimization design of
HarmTrace rather than from applying RL alone, we compare HarmTrace with PPO,
GRPO, DAPO, and PAPO under the same E-SFT initialization. As shown in
Tables~\ref{tab:rl-baselines-label-caa} and
\ref{tab:rl-baselines-target-entity-span-bbox}, the generic RL methods show
different strengths across metrics, while HarmTrace achieves the highest
scores on all reported harmfulness and target-identification metrics. Among
the generic RL methods, GRPO obtains the highest JRA of 49.09\%, whereas
HarmTrace reaches 52.51\%. Overall, the optimization design
of HarmTrace yields additional gains beyond those obtained by generic RL
under the same E-SFT initialization.

\begin{table}[!t]
\centering
\begin{tabular*}{\columnwidth}{@{\extracolsep{\fill}}l*{3}{c}@{}}
\toprule
\multirow{2}{*}{\textbf{Method}} &
\multicolumn{2}{c}{\textbf{Label}} &
\multirow{2}{*}{\textbf{JRA}} \\
\cmidrule(lr){2-3}
& Acc. & F1 & \\
\midrule
PPO & 77.61 & 72.33 & 46.80 \\
GRPO & \underline{78.74} & 74.49 & \underline{49.09} \\
DAPO & 78.08 & 72.94 & 47.03 \\
PAPO & 78.46 & \underline{74.61} & 48.40 \\
HarmTrace & \textbf{80.15} & \textbf{76.77} & \textbf{52.51} \\
\bottomrule
\end{tabular*}
\caption{Performance comparison of different RL methods under the same E-SFT-initialized Qwen3-VL-8B backbone.}
\label{tab:rl-baselines-label-caa}
\end{table}

\begin{table}[!t]
\centering
\small
\begin{tabular}{@{}l@{\hspace{0.35em}}c@{\hspace{0.35em}}c@{\hspace{0.35em}}c@{\hspace{0.35em}}c@{\hspace{0.35em}}c@{\hspace{0.35em}}c@{}}
\toprule
\multirow{2}{*}{\textbf{Method}} &
\multicolumn{1}{c}{\textbf{Category}} &
\multicolumn{1}{c}{\textbf{Entity}} &
\multicolumn{2}{c}{\textbf{Text}} &
\multicolumn{2}{c}{\textbf{Visual}} \\
\cmidrule{2-2}
\cmidrule{3-3}
\cmidrule(lr){4-5}
\cmidrule(lr){6-7}
& EM & $\mathrm{F1}_{\mathrm{rel}}$ & EM & F1 & IoU50 & IoU75 \\
\midrule
PPO & 63.93 & 61.42 & 54.57 & 58.09 & 58.68 & 55.02 \\
GRPO & 68.72 & 65.98 & \underline{57.99} & \underline{61.70} & 59.82 & 56.39 \\
DAPO & 65.30 & 61.19 & 55.02 & 59.10 & 60.27 & 57.08 \\
PAPO & \underline{70.55} & \underline{66.21} & 57.76 & 61.12 & \underline{63.93} & \underline{60.73} \\
HarmTrace & \textbf{72.60} & \textbf{68.72} & \textbf{60.73} & \textbf{64.93} & \textbf{66.67} & \textbf{62.56} \\
\bottomrule
\end{tabular}
\caption{Performance comparison of RL methods on
target-identification fields under the same E-SFT initialization.}
\label{tab:rl-baselines-target-entity-span-bbox}
\end{table}

\section{Conclusion}

We study fine-grained target identification in harmful memes, where models
jointly predict harmfulness, target category, target entity, textual mention,
and visual region. Together, these outputs form a structured record that can
support moderation review. We introduce Meme3W with unified, human-verified
annotations and Joint Record Accuracy (JRA) for strict record-level evaluation.
We further propose HarmTrace, an anchor-calibrated decoupled optimization
framework that combines entity-aware supervision, decoupled credit assignment,
and a Virtual Positive Anchor. Experimental results show that HarmTrace improves
JRA and all reported fine-grained target-identification fields. Future work will
evaluate HarmTrace on larger MLLMs.

\bibliography{aaai2027}

% Optional appendix / supplementary material. Comment out this line for the
% main-paper-only submission.

\section{A Meme3W Construction and Validation}
\label{app:dataset-construction-validation}
\suppressfloats[t]

This section details the construction and validation of Meme3W, including data collection and filtering, the annotation pipeline, annotation quality assessment, and dataset release.

\noindent\textbf{Disclaimer.} This paper contains harmful content, which has the potential to be offensive and may disturb readers.

\subsection{A.1 Data Curation, Unified Schema, and Annotation Prompt}
\label{app:data-schema-prompt}

\paragraph{Data sources and screening.}
Meme3W is curated from four existing multimodal meme datasets, namely PrideMM~\cite{shah2024memeclip}, MAMI~\cite{fersini2022semeval}, Hateful Memes (FHM)~\cite{kiela2020hateful}, and Harm-C~\cite{pramanick2021momenta}.
These datasets provide harmful and non-harmful examples, but their original
annotations cover only subsets of the target-identification fields and do not
jointly identify the attacked target and its textual and visual grounding.
FHM covers all target categories considered in Meme3W, although
racial, ethnic, and religious targets dominate its harmful subset. To broaden
and balance target coverage, we supplement it with harmful memes from MAMI,
PrideMM, and Harm-C, which focus on misogyny, LGBTQ-related harm, and
COVID-19-related harm, respectively. After removing samples with unclear or
unreliable visual content, we re-annotate the retained harmful memes under the
unified Meme3W schema. Table~\ref{tab:source-data} compares the annotation
fields provided by the source datasets with those defined in Meme3W.

\begin{table*}[t]
\centering
\small
\begin{tabular*}{\textwidth}{@{\extracolsep{\fill}}llrrccccc@{}}
\toprule
& & & & \multicolumn{5}{c}{Original annotations} \\
\cmidrule(lr){5-9}
Dataset & Topic & \#Total & \#Harm. & Harm. & Tgt. cat. & Tgt. ent. & Text & Box \\
\midrule
PrideMM~\cite{shah2024memeclip}
& LGBTQ+ & 5,063 & 2,482 & \cmark & \cmark & -- & -- & -- \\
MAMI~\cite{fersini2022semeval}
& Misogyny & 11,000 & 5,500 & \cmark & \cmark & -- & -- & -- \\
Hateful Memes~\cite{kiela2020hateful}
& General hate & 10,000 & 5,000 & \cmark & -- & -- & -- & -- \\
Harm-C~\cite{pramanick2021momenta}
& COVID-19 & 3,544 & 1,249 & \cmark & \cmark & -- & -- & -- \\
\midrule
\textbf{Meme3W}
& Mixed & \textbf{10,662} & \textbf{4,418} & \cmark & \cmark & \cmark & \cmark & \cmark \\
\bottomrule
\end{tabular*}
\caption{Annotation fields in the source datasets and Meme3W. Harm. denotes harmfulness labels. Tgt. cat. denotes source-specific target-category annotations defined under different taxonomies. Tgt. ent., Text, and Box denote target entities, textual mentions, and visual bounding boxes, respectively. A check mark indicates field availability.}
\label{tab:source-data}
\end{table*}

\paragraph{FHM category distribution before augmentation.}
Table~\ref{tab:fhm-only-category-dist} reports the target-category distribution
of the retained FHM harmful subset, which contains 3,548 memes after applying
the same filtering and deduplication criteria used for Meme3W.

\begin{table}[t]
\centering
\small
\begin{tabular*}{\columnwidth}{@{\extracolsep{\fill}}lrr@{}}
\toprule
Attack target & Count & Share (\%) \\
\midrule
Group background & 1,463 & 41.23 \\
Religion and caste & 1,090 & 30.72 \\
Gender & 677 & 19.08 \\
Health condition & 318 & 8.96 \\
\midrule
\textbf{Total} & \textbf{3,548} & \textbf{100.00} \\
\bottomrule
\end{tabular*}
\caption{Attack-target distribution of the retained FHM harmful subset before
augmentation with MAMI, PrideMM, and Harm-C. Counts use the same filtering
criteria as the final Meme3W dataset.}
\label{tab:fhm-only-category-dist}
\end{table}

\paragraph{Harmfulness definition.}
A meme is considered harmful when it directs harmful behavior toward a clear
target. Harmful behavior includes targeted satire, insults, mockery,
degradation, stereotyping, dehumanization, threats, exclusion, segregation,
discrimination, assertions of inferiority, comparisons to animals or objects,
mockery of hate crimes or historical suffering, and gender- or
sexuality-based harm. Ordinary humor, non-targeted profanity, general
criticism, and attacks on criminals, terrorists, or criminal activities are
considered non-harmful.

\paragraph{Unified schema and taxonomy.}
Each meme is annotated with a harmfulness label, target category, target
entity, textual mention, and visual region. Because non-harmful memes contain no attacked target, all target-identification fields other than the visual-region field are set to \texttt{null}, while the visual-region field is set to an empty list \texttt{[]}.
For harmful memes, the target category and target entity are always annotated. The textual mention and visual region are included only when corresponding evidence is available. Otherwise, the textual mention is set to \texttt{null}, and the visual region is set to \texttt{[]}.
A prediction is considered correct for an absent field only when
it likewise indicates the field's absence. The unified target-category
taxonomy contains four categories:

\begin{itemize}
    \item \textbf{group background} includes ethnicity,
    race, nationality, regional or ethnic origin, minority background, and
    immigration or migrant status.
    \item \textbf{religion and caste} includes religion,
    religious belief, religious identity, followers or believers of a
    religion, religious people, religious groups, and caste.
    \item \textbf{gender} includes biological sex, gender identity,
    and sexual orientation.
    \item \textbf{health condition} includes disability,
    disease, vaccination-related targets, infection, bodily condition, medical
    treatment, public-health measures, and health-related policy framing.
\end{itemize}

\paragraph{Unified output schema.}
Each complete model response operationalizes the explain-then-answer sequence
in Eq.~(2) of the main paper as a single flat JSON object. The
\texttt{think} field contains auxiliary reasoning. The remaining five fields
are \texttt{label}, \texttt{target\_category}, \texttt{target\_entity},
\texttt{text\_mention}, and \texttt{visual\_region}. Together, these five
fields form the structured prediction $a_i$ and follow the structured
annotation defined in Eq.~(1) of the main paper. In Eq.~(2),
$\mathrm{answer}:a_i$ denotes these five fields collectively rather than a
literal \texttt{answer} key. Only these five fields are used for semantic
matching and target-reward computation, while \texttt{think} is not
semantically scored. Before scoring, the response parser extracts them as the
input to the verifier. Field applicability and absence values follow the
unified schema defined above.

\paragraph{Shared prompting protocol.}
The three MLLMs used for candidate generation and the MLLMs evaluated in our
experiments follow the same task definition, five-step analysis procedure, and
structured output schema, with model-specific conversation templates applied
when necessary. The operational prompt instructions are provided below.

\begin{enumerate}
    \item \textbf{Analyze the visual and textual content.}
    First enumerate all important visible elements in the image, rather than
    considering only the most salient one. These may include every visible
    person or group, object, symbol, animal, scene, gesture, item of clothing,
    flag, religious marker, medical marker, and visible text region. Do not omit an element solely because it may not be the attacked target. Identify the element first and determine its role afterward. Also identify the relevant cues in the accompanying text.

    \item \textbf{Decode the meme semantics.}
    Analyze relevant linguistic and contextual cues, including puns, double
    meanings, metaphors, euphemisms, coded expressions, offensive terms,
    symbolic associations, stereotypes, and historical, cultural, or social
    context. Consider whether the meme conveys insults, discrimination,
    dehumanization, exclusion, inferiority, or mockery of hate crimes,
    historical persecution, violence, disability, disease, or suffering. If
    no such semantic cue is relevant, state this briefly.

    \item \textbf{Distinguish the attacked target from contextual elements.}
    Identify visually salient figures and symbols, and determine whether each
    is itself being attacked or instead functions as a perpetrator, criminal,
    terrorist, background object, metaphor, symbol, or visual vehicle used to
    attack another target. Identify the actual attacked target based on the
    complete meaning of the meme.

    \item \textbf{Determine the target evidence.}
    For a harmful meme, record the exact textual mention only when the attacked
    target is explicitly referred to in the accompanying text, and identify the
    corresponding visual region or regions only when the attacked target is
    explicitly depicted in the image. Otherwise, assign the predefined absence
    value to the corresponding field.

    \item \textbf{Classify the meme and target category.}
    Explicitly determine whether harmful behavior is directed toward a clear
    target and identify the behavior type that is present or absent. Relevant
    behavior types include targeted satire, insult, mockery, degradation,
    stereotyping, dehumanization, threats, exclusion, segregation,
    discrimination, and assertions of inferiority. If no harmful behavior is
    directed toward a clear target, classify the meme as non-harmful. For a
    harmful meme, determine the target category and target entity. Finally,
    return the result using the unified output schema.

\end{enumerate}

\paragraph{Output format and normalization.}
Candidate models receive the meme image, its associated text, the task
instructions, and the unified output schema. Each complete response contains
intermediate analysis in \texttt{think} together with the five prediction
fields in a single flat JSON object. The \texttt{text\_mention} field contains
the exact text span referring to the attacked target, while the
\texttt{visual\_region} field contains a list of target bounding boxes, each represented as
$[x_1,y_1,x_2,y_2]$. All bounding-box coordinates are normalized to the range
$[0,1000]$, where $(x_1,y_1)$ and $(x_2,y_2)$ denote the top-left and
bottom-right corners, respectively. The parser extracts the five prediction
fields, and the verifier validates their values. The resulting
predictions are presented to annotators as editable references. Field absence
follows the unified schema defined above.

\subsection{A.2 Human Annotation and Quality Control}
\label{app:annotation-quality-control}

\paragraph{Annotation guidelines.}
Annotators follow the requirements below:

\begin{enumerate}
    \item \textbf{Determine harmfulness and target category.}
    Annotators first determine whether the meme is harmful. For each harmful
    meme, the attacked target is assigned to exactly one category from the
    taxonomy defined in Appendix~A.1. Samples whose target category cannot be
    determined reliably, or that independently attack multiple categories,
    are excluded rather than forced into a single category.

    \item \textbf{Interpret the meme from the author's perspective.}
    Harmfulness and target identity are determined from the intended meaning
    conveyed by the meme author, based only on the image--text content.
    Annotators consider satire, metaphor, insinuation, double meanings,
    stereotypes, and relevant historical, cultural, or social context rather
    than relying only on the literal wording.

    \item \textbf{Identify the actual attacked entity.}
    The target entity must precisely describe the person or group being
    attacked at an appropriate level of granularity. Visually salient
    perpetrators, symbols, background figures, and contextual objects must not
    be annotated as targets unless they are themselves being attacked. When
    the target is conveyed indirectly, the entity is inferred from the
    complete meaning of the meme.

    \item \textbf{Copy textual mentions exactly.}
    When the attacked target is explicitly referred to in the associated text,
    annotators copy the exact text span without paraphrasing, expanding, or
    correcting it. If no explicit textual mention is present, the field is set
    to \texttt{null}.

    \item \textbf{Mark complete visual target regions.}
    When the attacked target is visually depicted, each bounding box should
    tightly cover the complete visible extent of the target rather than only
    its face or another salient part. Multiple boxes are used when multiple
    visual instances of the same target are present. If the target is not
    explicitly depicted, the visual-region field is set to \texttt{[]}.

    \item \textbf{Ensure accurate multimodal grounding.}
    Textual mentions and visual regions must accurately correspond to the
    attacked target in the respective modality.
\end{enumerate}

\begin{figure}[t]
\centering
\includegraphics[width=\columnwidth]{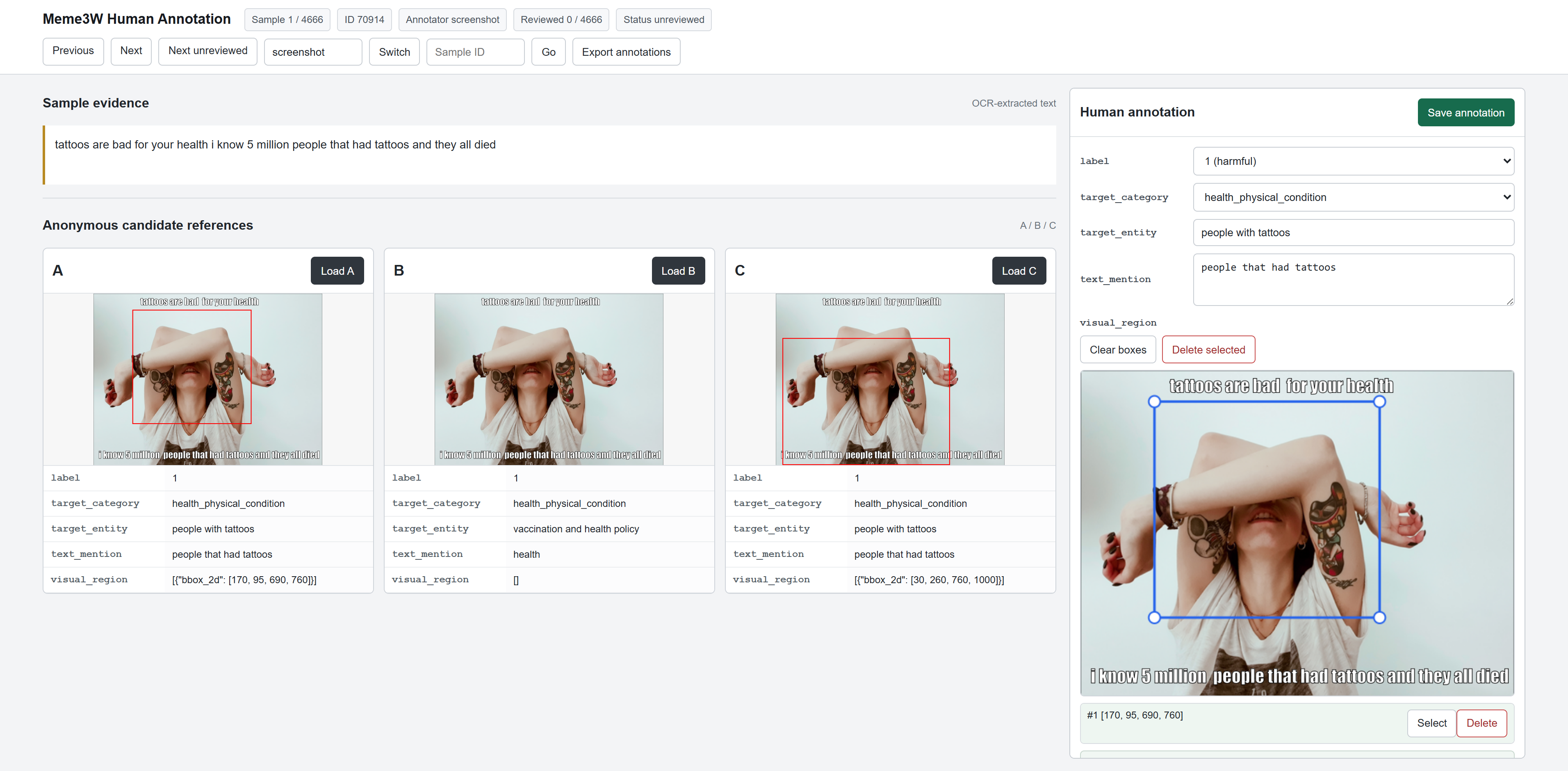}
\caption{Candidate-assisted human-annotation interface. Annotators inspect
the source text and three anonymized MLLM candidates, revise all structured
fields, and draw or adjust image-target boxes before saving the annotation.}
\label{fig:app-candidate-assisted-ui}
\end{figure}

\paragraph{Independent annotation and adjudication.}
Each meme is independently annotated by two annotators using editable MLLM
candidates as references. The identity of the model producing each candidate
is hidden, and candidate positions are randomly ordered in the annotation
interface. Disagreements are independently reviewed by a third annotator and
then resolved through discussion to produce the final gold annotation.
Figure~\ref{fig:app-candidate-assisted-ui} shows the interfaces used for the
initial independent annotation and the third-annotator review.

\paragraph{Entity canonicalization.}
Before computing target-entity agreement and evaluation metrics, we apply a deterministic canonicalization procedure to normalize surface-form variation. We normalize casing, punctuation, and whitespace, and map referentially equivalent spelling, plurality, abbreviation, and referring-expression variants to a shared canonical form. Broader groups and their subgroups, as well as modifiers that alter the target identity, remain distinct. The canonicalization lexicon was manually reviewed, finalized prior to model evaluation, and kept fixed across all experiments. The canonicalized mentions are subsequently compared using $F1_{\mathrm{rel}}$.
For JRA, the target-entity field is considered correct when
$F1_{\mathrm{rel}} \geq 0.7$.

\begin{table}[t]
\centering
\small
\begin{tabular*}{\columnwidth}{@{\extracolsep{\fill}}p{0.31\columnwidth}p{0.34\columnwidth}p{0.18\columnwidth}@{}}
\toprule
Entity A & Entity B & Same Form \\
\midrule
women & woman & Yes \\
Muslim people & Muslims & Yes \\
Jewish people & Jews & Yes \\
LGBTQ people & LGBTQ community & Yes \\
immigrants & refugees & No \\
Asian people & Chinese people & No \\
Black women & women & No \\
\bottomrule
\end{tabular*}
\caption{Examples of target-entity canonicalization. \emph{Yes} indicates that two mentions map to the same canonical form; \emph{No} indicates that they remain distinct. Final target-entity correctness is evaluated using $F1_{\mathrm{rel}}$.}
\label{tab:app-entity-canonicalization-examples}
\end{table}

\begin{figure}[t]
\centering
\includegraphics[width=\columnwidth]{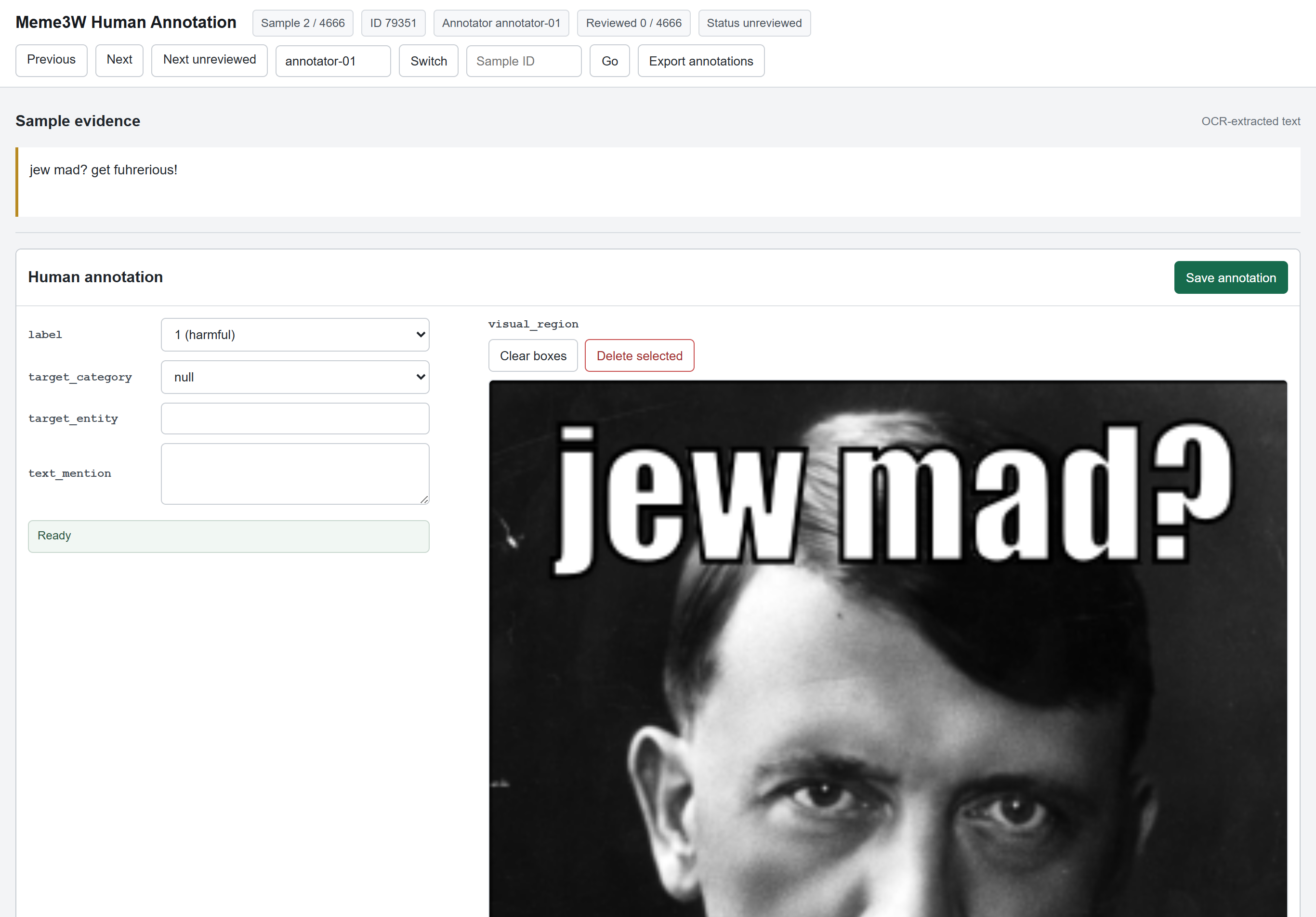}
\caption{Candidate-blind annotation interface without access to MLLM candidates.}
\label{fig:app-candidate-blind-ui}
\end{figure}

\paragraph{Candidate-blind audit.}
We conduct a candidate-blind audit on 200 cases sampled to cover all four
target categories and different textual and visual grounding conditions.
Annotators receive only the raw meme and annotation guidelines, without
access to the MLLM candidate annotations. The blind annotations are compared
with the final gold annotations using the same field-specific metrics as in
the main annotation process. Figure~\ref{fig:app-candidate-blind-ui} shows the
candidate-blind annotation interface.

\begin{table}[t]
\centering
\begin{tabular*}{\columnwidth}{@{\extracolsep{\fill}}llc@{}}
\toprule
Field & Metric & Blind vs. Gold \\
\midrule
Target category & Category agreement & 0.900 \\
Target entity & Normalized F1 & 0.797 \\
Textual mention & Token-F1 & 0.808 \\
Visual region & IoU & 0.803 \\
\textbf{Unweighted mean} & -- & \textbf{0.827} \\
\bottomrule
\end{tabular*}
\caption{Candidate-blind audit against the final gold annotations.}
\label{tab:app-blind-audit}
\end{table}

On this subset, re-evaluation against the blind annotations preserved the
relative model ordering obtained with the final gold annotations, suggesting
that candidate assistance did not alter the main comparative conclusion.

\subsection{A.3 Dataset Statistics and Split Integrity}
\label{app:dataset-statistics}

\begin{figure}[t]
\centering
\includegraphics[width=\columnwidth]{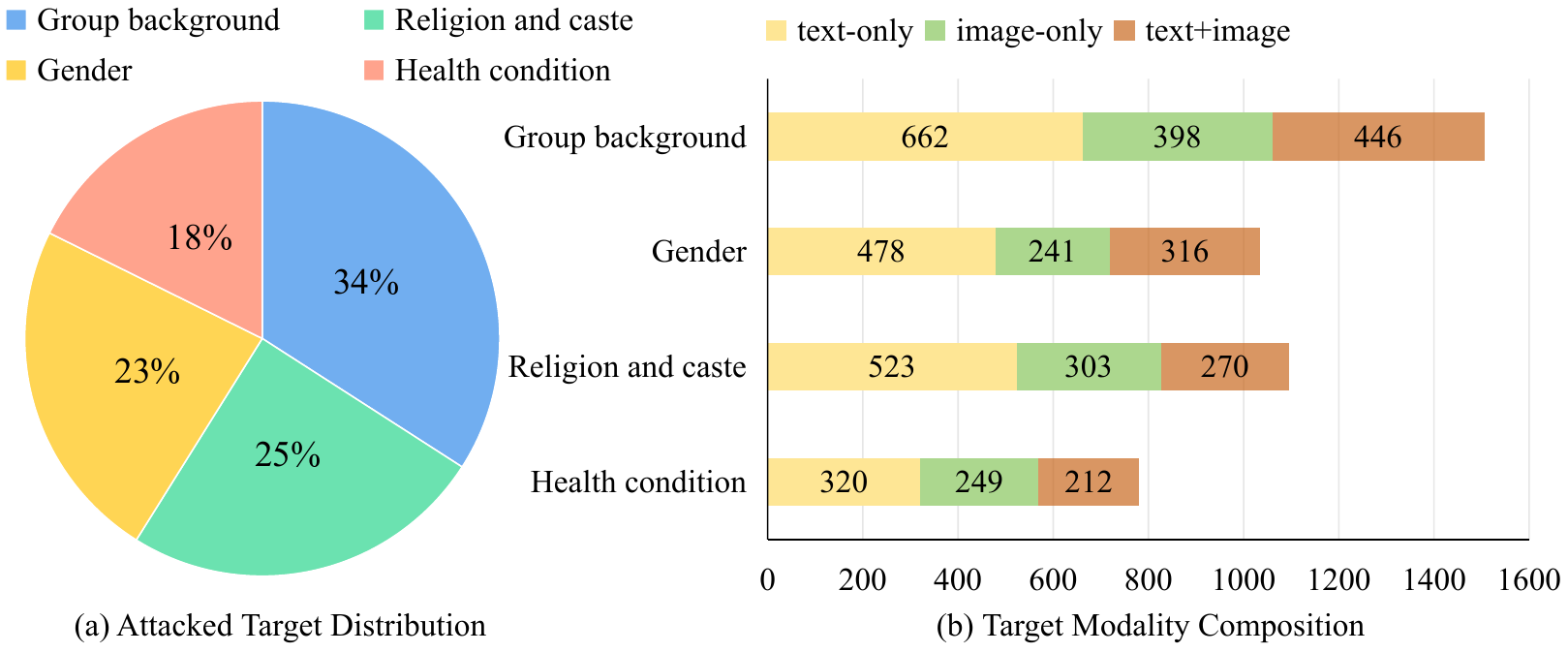}
\caption{Attack-target distribution and target-modality composition among the
4,418 harmful memes in Meme3W.}
\label{fig:app-dataset-category}
\end{figure}

\paragraph{Dataset statistics.}
Meme3W contains 10,662 memes, including 4,418 harmful and 6,244
non-harmful examples. Table~\ref{tab:app-split-statistics} reports the
exact numbers of examples in each split. Figure~\ref{fig:app-dataset-category}
shows the overall attack-target distribution and target-modality composition
of the harmful subset.

\begin{table}[t]
\centering
\small
\begin{tabular*}{\columnwidth}{@{\extracolsep{\fill}}lrrr@{}}
\toprule
Split & Harmful & Non-harmful & Total \\
\midrule
Training   & 3,759 & 5,307 & 9,066 \\
Validation & 221   & 312   & 533 \\
Test       & 438   & 625   & 1,063 \\
\midrule
Total      & 4,418 & 6,244 & 10,662 \\
\bottomrule
\end{tabular*}
\caption{Numbers of harmful and non-harmful memes in each Meme3W split.}
\label{tab:app-split-statistics}
\end{table}

\paragraph{Split integrity.}
We verify that no exact image duplicates occur across the training,
validation, and test splits using pixel-level hashing. Near-duplicate
candidates identified through perceptual hashing are manually reviewed, and
verified near-duplicate groups are assigned to the same split. Recurring meme
templates with different textual or visual content are retained as distinct
samples.

\paragraph{Data release and responsible use.}
We will release the structured annotations, fixed data splits, annotation
guidelines, shared prompt, and evaluation code. Raw images or source
identifiers will be distributed in accordance with the licenses of the
corresponding source datasets. Meme3W contains offensive and discriminatory
content and is intended for research on harmful-content understanding and
moderation.

\section{B Experimental Details and Implementation}
\label{app:implementation-details}

\subsection{B.1 Implementation and Evaluation Protocol}
\label{app:evaluation-protocol}

\paragraph{Shared evaluation prompt.}
All evaluated MLLMs use the shared task instructions, five-step analysis
procedure, and structured output schema described in Appendix~A. Each response
follows the explain-then-answer format. The \texttt{think} field contains
intermediate analysis, while only the five prediction fields are used for
evaluation.

\paragraph{Model and decoding settings.}
For API-based baselines, we use the official model endpoints available at the
time of evaluation. For open-source baselines, we use the publicly released
instruction-tuned versions of the models reported in the main-results table. All models follow
the same task definition, output schema, and parsing rules. Malformed
structured outputs are treated as invalid predictions. ExPO-HM follows an SFT-to-RL pipeline. Because its task-specific SFT data are not publicly available, we initialize it from our standard SFT checkpoint trained on Meme3W and subsequently apply the ExPO-HM optimization procedure.

\paragraph{Training hyperparameters.}
We use ms-swift for supervised fine-tuning and verl for reinforcement learning.
All trainable methods use LoRA \cite{Hu2021LoRALA} with rank $64$ and $\alpha=128$. The key
hyperparameters are summarized in Tables~\ref{tab:app-sft-hyperparameters}
and~\ref{tab:app-rl-hyperparameters}. All hyperparameters and checkpoint-selection criteria for the reported systems were determined exclusively on the validation split. Test-set results were computed only after the corresponding configurations had been fixed, and no test result was used for hyperparameter or checkpoint selection.

\begin{table}[t]
\centering
\begin{tabular*}{\columnwidth}{@{\extracolsep{\fill}}lc@{}}
\toprule
Parameter & Value \\
\midrule
LoRA rank / alpha & 64 / 128 \\
Epochs & 3 \\
Learning rate & $1\times10^{-4}$ \\
Max sequence length & 4096 \\
Per-device train batch size & 8 \\
Gradient accumulation steps & 2 \\
Effective batch size & 32 \\
\bottomrule
\end{tabular*}
\caption{SFT hyperparameter settings.}
\label{tab:app-sft-hyperparameters}
\end{table}

\begin{table}[t]
\centering
\begin{tabular*}{\columnwidth}{@{\extracolsep{\fill}}lc@{}}
\toprule
Parameter & Value \\
\midrule
LoRA rank / alpha & 64 / 128 \\
Epochs & 3 \\
Train batch size & 32 \\
PPO mini-batch size & 32 \\
Rollout group size & 8 \\
Clip ratio & 0.2 \\
KL coefficient & 0.02 \\
Rollout temperature & 0.8 \\
\bottomrule
\end{tabular*}
\caption{RL hyperparameter settings.}
\label{tab:app-rl-hyperparameters}
\end{table}

\subsection{B.2 Teacher-Trajectory Generation and Filtering}
\label{app:teacher-filtering}

We use Gemini 3 Flash to generate one explain-then-answer trajectory for each
of the 9,066 training samples. Each trajectory first provides intermediate
analysis in the \texttt{think} field and then produces the five prediction
fields in the same flat JSON object. After automatic filtering, 6,005
trajectories are retained. A trajectory is retained only when its response is
parseable and all five fields satisfy the corresponding gold-matching
requirements.

The filtered trajectories are used only for cold-start SFT and E-SFT.
Standard SFT and E-SFT use the same filtered trajectory set and differ only in
token weighting. The trajectory filtering does not remove samples from subsequent policy
optimization. The subsequent policy-optimization stage uses the full training
set.

\subsection{B.3 Reward Components and Active Target-Identification Fields}
\label{app:reward-components}

\paragraph{Format and label reward.}
Following Eq.~(1) of the main paper, only the five prediction fields in $a_i$
enter semantic reward computation. The response parser extracts these fields
from the flat response object. The verifier then checks the resulting five-field
JSON representation against the unified output schema defined in Appendix~A.1
and validates the prediction-field value types and bounding-box coordinates.
Invalid JSON, missing or additional prediction fields, invalid
field values, and invalid bounding boxes fail the format check. The binary label reward is
\begin{equation}
r_i^{\mathrm{lab}}
=
\mathbf{1}[\mathrm{format\ valid}]
\mathbf{1}[\hat l_i=l_i^\star].
\end{equation}
Responses that fail the format check receive zero label reward. For a
non-harmful prediction, \texttt{target\_category},
\texttt{target\_entity}, and \texttt{text\_mention} must be
\texttt{null}, while \texttt{visual\_region} must be \texttt{[]}.
Target-identification rewards are computed only for format-valid,
label-correct responses to gold-harmful samples.

\paragraph{Applicable-field reward.}
The target-identification reward provides graded credit for partially correct
predictions while preserving complete-record consistency. For a gold-harmful
sample, let $c$, $e$, $s$, $B$, and $q$ denote target category, target entity,
textual mention, visual region, and schema consistency, respectively. The
active component set is
\begin{equation}
\mathcal{A}_i
=
\{c,e,q\}
\cup
\{s:s_i^\star\neq\texttt{null}\}
\cup
\{B:B_i^\star\neq\emptyset\}.
\end{equation}
Category, entity, and schema consistency are always active. Textual and visual
components are included only when the corresponding gold evidence is present.
The active scores are combined as
\begin{equation}
r_i^{\mathrm{tag}}
=
\frac{\sum_{f\in\mathcal{A}_i}w_f r_{i,f}}
{\sum_{f\in\mathcal{A}_i}w_f},
\end{equation}
where the fixed weights are $0.25$ for category, $0.25$ for entity, $0.20$ for
textual mention, $0.20$ for visual region, and $0.10$ for schema consistency.
These weights sum to 1. When textual
or visual evidence is absent, the denominator
renormalizes the remaining active weights. Therefore,
$r_i^{\mathrm{tag}}\in[0,1]$ and reaches 1 when all active components
receive a score of 1.

The category component uses exact match. The entity component is
\begin{equation}
r_{i,e}
=
\max\!\left(
S_{\mathrm{rel}},
\lambda\,\mathrm{F1}_{\mathrm{ent}}
+
(1-\lambda)\,\mathrm{EM}_{\mathrm{ent}}
\right),
\end{equation}
where $\lambda=0.7$, and $S_{\mathrm{rel}}$ follows the entity canonicalization
and relation matching described in Appendix~A.2. The textual component is
\begin{equation}
r_{i,s}
=
\max\!\left(
\mathrm{EM}_{\mathrm{tok}},
\lambda\,\mathrm{F1}_{\mathrm{tok}}
+
(1-\lambda)\,\mathrm{EM}_{\mathrm{tok}}
\right).
\end{equation}
The schema-consistency component equals 1 only when the predicted category is
valid and the predicted presence or absence of both textual and visual
evidence matches the gold annotation. Otherwise, it equals 0.

\paragraph{Visual-region reward and penalties.}
The visual component uses soft matching to provide graded credit for spatially
close predictions, including imperfect box matches.
For a predicted box $b$ and gold box $g$, the pairwise score is
\begin{equation}
S_B(b,g)
=
\max\!\left\{
\begin{aligned}
&\mathrm{IoU},\\
&\gamma_{\mathrm{IoU}}
\min\!\left(1,\frac{\mathrm{IoU}}{\tau_{\mathrm{soft}}}\right)
+\gamma_{\mathrm{Cov}}\,\mathrm{Cov}\\
&\qquad
+\gamma_{\mathrm{Prec}}\,\mathrm{Prec}
+\gamma_{\mathrm{ctr}}\,S_{\mathrm{ctr}}
\end{aligned}
\right\},
\end{equation}
where $(\gamma_{\mathrm{IoU}},\gamma_{\mathrm{Cov}},
\gamma_{\mathrm{Prec}},\gamma_{\mathrm{ctr}})
=(0.55,0.20,0.15,0.10)$ and $\tau_{\mathrm{soft}}=0.6$.
Here, $\mathrm{Cov}$, $\mathrm{Prec}$, and $S_{\mathrm{ctr}}$ denote
gold-region coverage, prediction precision, and center proximity.
For multiple regions, predicted and gold boxes are greedily matched and
their scores are averaged. The minimum enclosing rectangle is also considered
when multiple predicted boxes jointly cover one gold region. Each additional
predicted box with best-match score below $0.20$ incurs a $0.05$ penalty.
If the gold annotation contains no visual region, the visual component is
inactive and any predicted region sets the schema-consistency score to 0.

\subsection{B.4 Threshold Sensitivity and Visual Matching Robustness}
\label{app:threshold-visual-robustness}

\begin{figure}[t]
\centering
\includegraphics[width=\columnwidth]{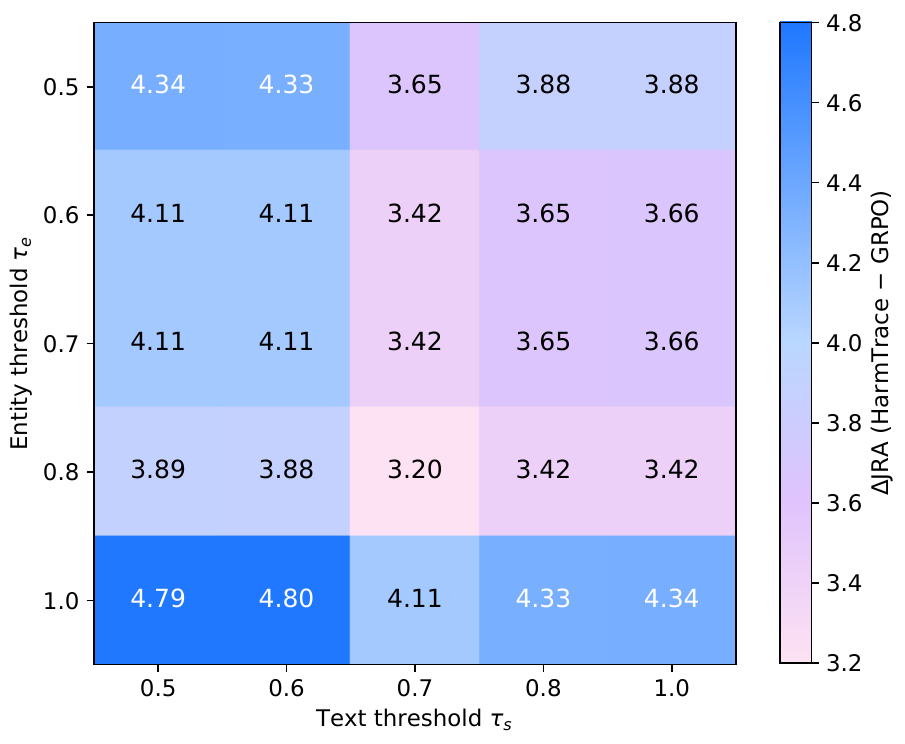}
\caption{Sensitivity of the JRA improvement to the target-entity threshold
$\tau_e$ and textual-mention threshold $\tau_s$. Each cell reports the JRA
difference between HarmTrace and GRPO in percentage points.}
\label{fig:jra-threshold-sensitivity}
\end{figure}

\paragraph{Entity and text threshold sensitivity.}
We set $\tau_e=\tau_s=0.7$ for target-entity and textual-mention matching.
To assess sensitivity to these choices, we vary $\tau_e$ and $\tau_s$ over
$\{0.5,0.6,0.7,0.8,1.0\}$ and evaluate JRA on the complete $5\times5$ grid.
Figure~\ref{fig:jra-threshold-sensitivity} shows the JRA difference between
HarmTrace and GRPO under each threshold combination. HarmTrace remains above
GRPO for all 25 settings, with $\Delta$JRA ranging from 3.20 to 4.80 points,
indicating that the model ordering is not sensitive to the selected entity or
textual-mention threshold.

\paragraph{Visual IoU threshold sensitivity.}
With $\tau_e=\tau_s=0.7$ fixed, we vary the minimum-enclosing-rectangle
IoU threshold. As shown in Table~\ref{tab:jra-iou-threshold-sensitivity},
HarmTrace remains above GRPO at every evaluated threshold.

\begin{table}[t]
\centering
\setlength{\tabcolsep}{2.5pt}
\begin{tabular*}{\columnwidth}{@{\extracolsep{\fill}}lrrrrr@{}}
\toprule
Model & IoU 0.5 & IoU 0.6 & IoU 0.7 & IoU 0.75 & IoU 0.8 \\
\midrule
GRPO & 49.09 & 47.49 & 46.35 & 46.35 & 44.52 \\
HarmTrace & 52.51 & 52.05 & 50.46 & 50.23 & 50.00 \\
$\Delta$ & +3.42 & +4.57 & +4.11 & +3.88 & +5.48 \\
\bottomrule
\end{tabular*}
\caption{Sensitivity of JRA to the minimum-enclosing-rectangle IoU threshold
with $\tau_e=\tau_s=0.7$. Values are percentages on the 438 gold-harmful test
memes, and $\Delta$ denotes HarmTrace minus GRPO.}
\label{tab:jra-iou-threshold-sensitivity}
\end{table}

\paragraph{Visual-region matching robustness.}
We evaluate whether the JRA comparison depends on the visual-region matching
rule. In addition to minimum-enclosing-rectangle IoU, we consider one-to-one
bipartite matching with no unmatched boxes and geometric set IoU over the
unions of predicted and gold regions. All rules use an IoU threshold of 0.5,
with the remaining JRA criteria unchanged. HarmTrace remains above GRPO under
all three matching rules, indicating that the model ordering is not determined
by the visual-region matching implementation.

\begin{table}[t]
\centering
\setlength{\tabcolsep}{4pt}
\begin{tabular*}{\columnwidth}{@{\extracolsep{\fill}}lrrr@{}}
\toprule
Visual rule & GRPO & HarmTrace & $\Delta$JRA \\
\midrule
Enclosing-rectangle IoU & 49.09 & 52.51 & +3.42 \\
Bipartite set matching & 46.12 & 50.00 & +3.88 \\
Geometric set IoU & 49.77 & 52.28 & +2.51 \\
\bottomrule
\end{tabular*}
\caption{JRA under alternative visual-region matching rules. HarmTrace remains
above GRPO under all three rules.}
\label{tab:jra-visual-matching}
\end{table}

\begin{table}[t]
\centering
\small
\setlength{\tabcolsep}{2.5pt}
\begin{tabular*}{\columnwidth}{@{\extracolsep{\fill}}lrrrr@{}}
\toprule
Comparator & JRA & HarmTrace & $\Delta$JRA & Paired 95\% CI \\
\midrule
E-SFT & $45.89$ & $52.51$ & $+6.62$ & $[2.51,10.96]$ \\
GRPO & $49.09$ & $52.51$ & $+3.42$ & $[0.23,6.85]$ \\
\bottomrule
\end{tabular*}
\caption{Paired comparisons of JRA on the 438 gold-harmful test memes.
All values are percentages. Confidence intervals are percentile intervals
from 10,000 paired bootstrap resamples of test instances.}
\label{tab:jra-significance}
\end{table}

\begin{table}[!t]
\centering
\small
\setlength{\tabcolsep}{1.5pt}
\begin{tabular*}{\columnwidth}{@{\extracolsep{\fill}}lrrrr@{}}
\toprule
\textbf{Cat.} & \textbf{\#} & \textbf{E-SFT} & \textbf{GRPO} & \textbf{Ours} \\
\midrule
Group/bg. & 146 & 46.58 & 50.00 & \textbf{56.16} \\
Religion/caste & 111 & 60.36 & \textbf{63.96} & 62.16 \\
Gender & 103 & 44.66 & 47.57 & \textbf{56.31} \\
Health cond. & 78 & 25.64 & \textbf{28.21} & 26.92 \\
\midrule
Overall & 438 & 45.89 & 49.09 & \textbf{52.51} \\
\bottomrule
\end{tabular*}
\caption{Category-wise JRA for E-SFT, GRPO, and HarmTrace on
gold-harmful test memes.}
\label{tab:category-wise-caa}
\end{table}

\subsection{B.5 Additional Ablations and Robustness Analyses}
\label{app:additional-ablations-plan}

We report supplementary statistical and robustness analyses that are not
included in the main paper.

\paragraph{Statistical significance analysis.}
We quantify uncertainty in the primary JRA metric using 10,000 paired
percentile-bootstrap resamples of the 438 gold-harmful test memes, applying
the same resampled indices to both systems in each comparison. As shown in
Table~\ref{tab:jra-significance}, HarmTrace achieves 52.51% JRA, improving
over E-SFT (45.89\%) by 6.62 percentage points, with a paired 95% confidence
interval of $[2.51,10.96]$. Under the same E-SFT initialization, it also
outperforms GRPO (49.09\%) by 3.42 points, with a paired 95\% confidence
interval of $[0.23,6.85]$.

\paragraph{Category-wise target identification.}
To examine whether the overall JRA gains extend across target categories, we
compare E-SFT, GRPO, and HarmTrace. Table~\ref{tab:category-wise-caa}
reports category-wise JRA on gold-harmful test memes using the same joint
correctness criteria as the overall evaluation.

HarmTrace achieves the highest overall JRA, outperforming E-SFT and GRPO by
6.62 and 3.42 points, respectively. Its gains are concentrated in
\emph{group background} and \emph{gender}, where it improves over GRPO by
6.16 and 8.74 points and over E-SFT by 9.58 and 11.65 points. On
\emph{religion and caste} and \emph{health condition}, HarmTrace remains above
E-SFT but is slightly below GRPO by 1.80 and 1.29 points. These differences
correspond to only two and one test examples, respectively. Overall, the gains
on group background and gender outweigh these minor decreases.

\section{C. Qualitative Case Studies}
\label{app:additional-results}

\begin{figure*}[p]
\centering
\begin{minipage}{0.90\textwidth}
\centering
\includegraphics[width=\linewidth]{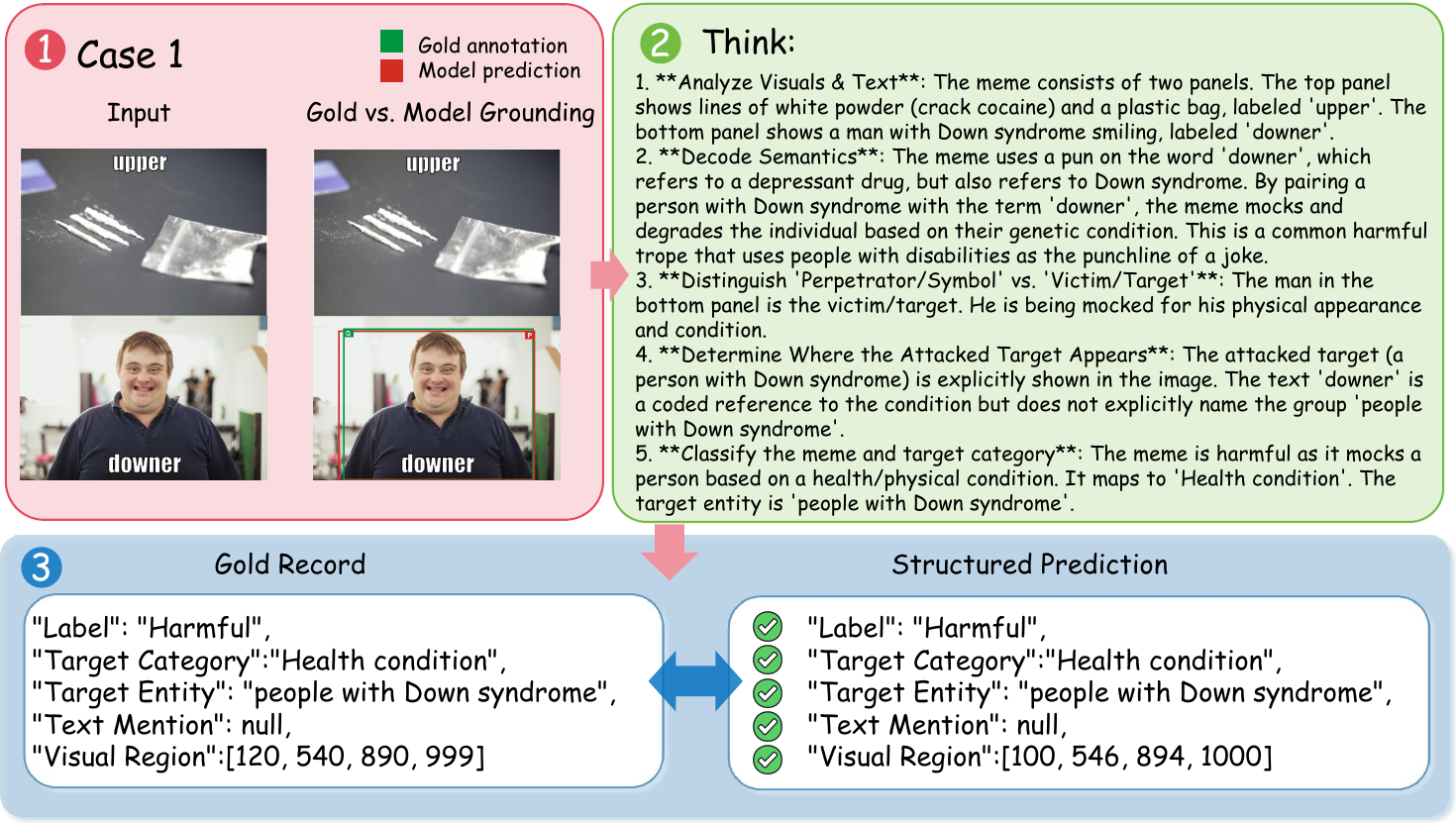}
\captionof{figure}{Case 1 demonstrates correct image-grounded target identification.}
\label{fig:case-study-1}
\end{minipage}

\par\medskip

\begin{minipage}{0.90\textwidth}
\centering
\includegraphics[width=\linewidth]{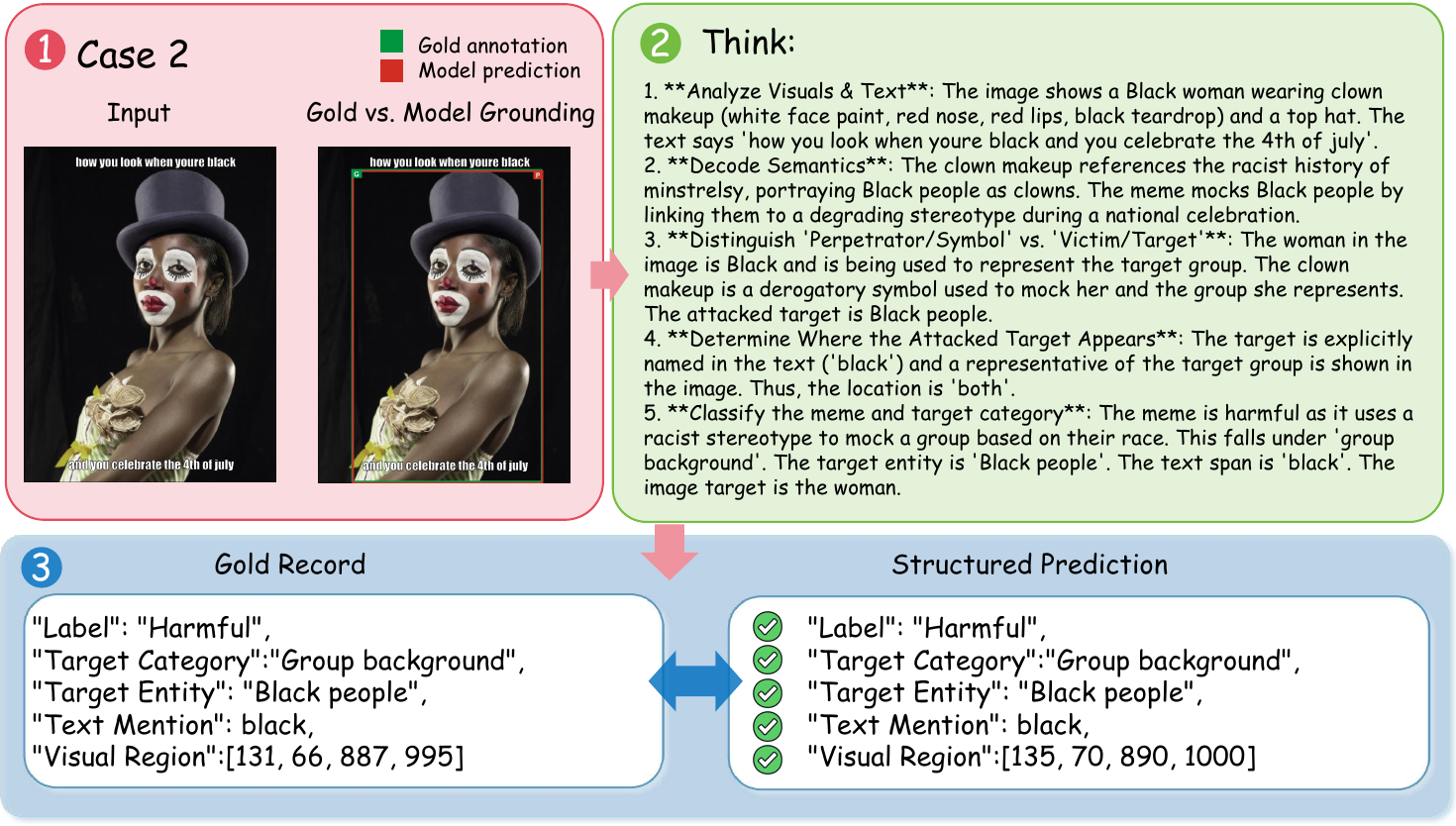}
\captionof{figure}{Case 2 demonstrates correct joint text--image target identification.}
\label{fig:case-study-2}
\end{minipage}
\end{figure*}

Figures~\ref{fig:case-study-1} and~\ref{fig:case-study-2} present two
representative successful cases. In both cases, HarmTrace correctly predicts
the harmfulness label and all target-identification fields. These fields
include the target category, target entity, textual mention, and visual region.
The resulting records therefore satisfy all JRA criteria.

\begin{figure*}[p]
\centering
\begin{minipage}{0.90\textwidth}
\centering
\includegraphics[width=\linewidth]{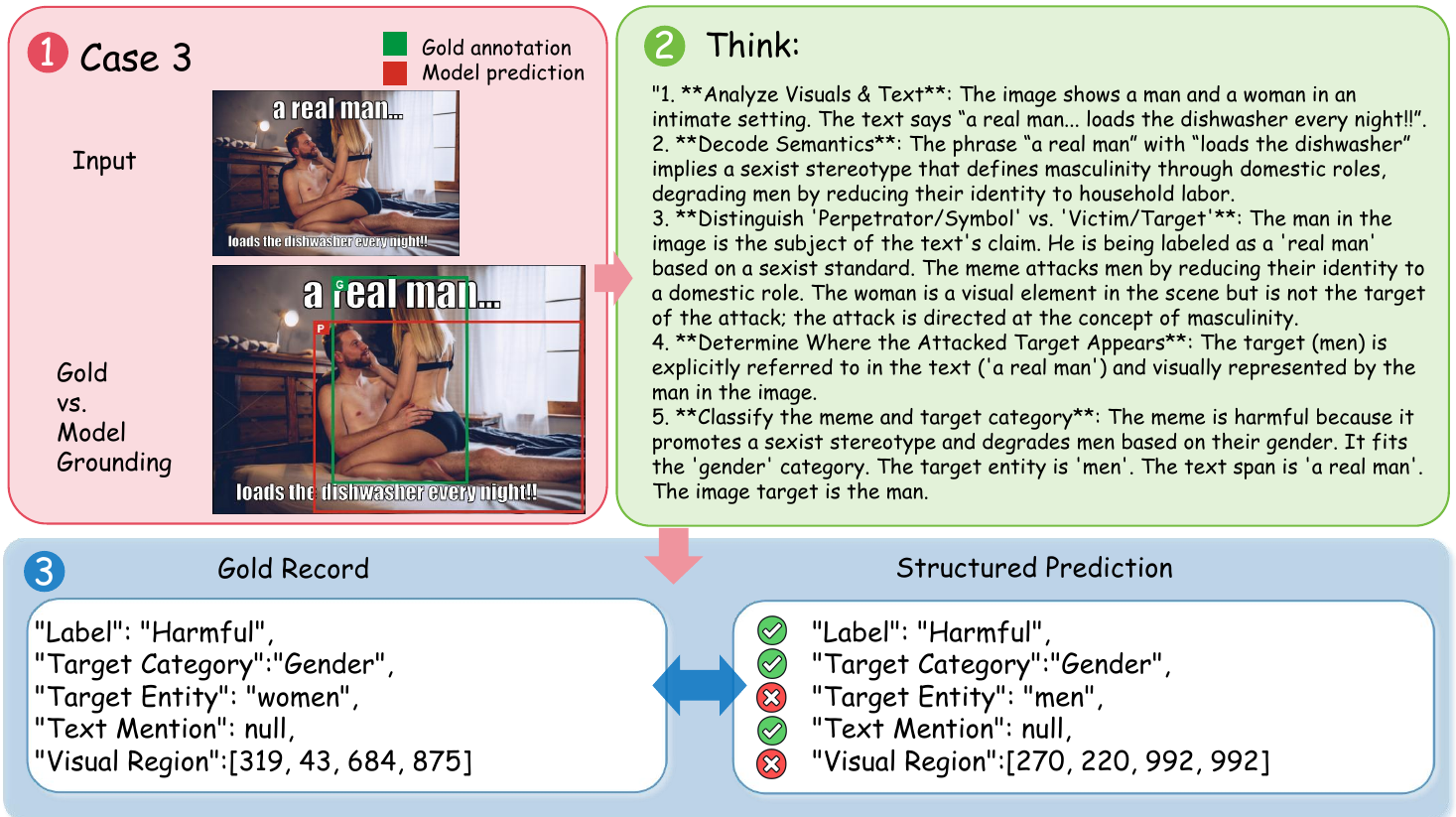}
\captionof{figure}{Case 3 shows target-entity and visual-grounding errors despite correct harmfulness and target-category predictions.}
\label{fig:case-study-3}
\end{minipage}

\par\medskip

\begin{minipage}{0.90\textwidth}
\centering
\includegraphics[width=\linewidth]{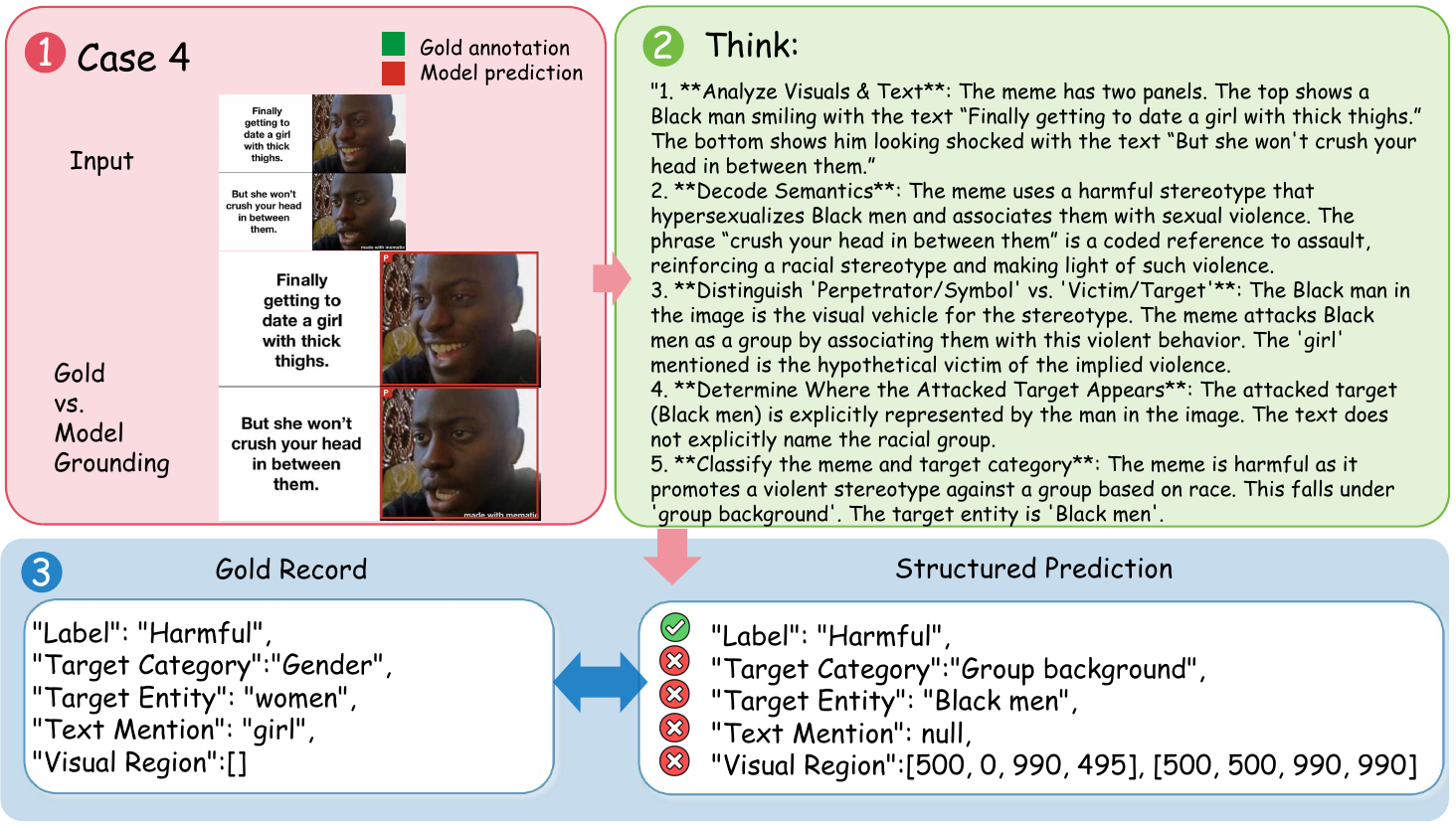}
\captionof{figure}{Case 4 shows cascading target-identification errors despite a correct harmfulness prediction.}
\label{fig:case-study-4}
\end{minipage}
\end{figure*}

Figures~\ref{fig:case-study-3} and~\ref{fig:case-study-4} illustrate two
different forms of cross-modal target misidentification. In the first case,
the text explicitly mentions a man. However, ``dishwasher'' is used as a
derogatory reference to the woman shown in the image. HarmTrace follows the
explicit mention of the man. It therefore fails to associate the implicit
textual attack with the woman. In the second case, the attacked target is a
girl mentioned only in the text. The image, however, depicts a Black man. HarmTrace relies on the visible person and incorrectly predicts and grounds the
man as the target. These cases suggest that the model may over-rely on a single
modality when textual and visual cues are not directly aligned. This can lead
to incorrect identification of the attacked target.

\section{D. Ethical Considerations}
\label{app:ethical-considerations}

\paragraph{Data use and release.}
Meme3W is derived from four publicly released research datasets, namely PrideMM, MAMI, Hateful Memes, and Harm-C. We will comply with their respective licenses and redistribution requirements. When raw-image redistribution is not
permitted, we will release only structured annotations, source identifiers,
and processing code. Because the data may contain identifiable individuals,
slurs, or stigmatizing content, the release will include a content warning and
a procedure for reviewing removal requests.

\paragraph{Annotator welfare and oversight.}
Five graduate-student annotators were informed in advance that the task
involved potentially offensive and discriminatory content. Participation was
voluntary, and annotators could skip individual examples or withdraw from the
task. They were compensated at a rate of \$10 per hour, consistent with
institutional requirements and above the applicable local minimum wage. MLLM
outputs were used only as editable annotation candidates. Disagreements were
resolved through independent human review.

\paragraph{Intended use.}
Meme3W is intended to support research on harmful-content understanding,
safety evaluation, model analysis, and content-moderation review. Its
structured annotations make the attacked target and the associated textual
and visual evidence explicit.


\end{document}